\documentclass[preprint,sort,compress,12pt]{elsarticle}

\usepackage{amssymb}
\usepackage{amsthm}
\usepackage{amsmath}
\usepackage{mathtools}
\usepackage{mathrsfs}
\usepackage{bm}
\usepackage{anyfontsize}
\usepackage{algorithm}
\usepackage[algo2e]{algorithm2e}
\usepackage{algpseudocode}
\usepackage{array}
\usepackage{multirow}
\usepackage{listings}
\usepackage{tabu}
\usepackage{enumerate}
\usepackage{fullpage}
\usepackage{xcolor}
\usepackage[colorinlistoftodos]{todonotes}
\usepackage{bbm}
\usepackage[colorlinks=true]{hyperref}
\usepackage{url}
\usepackage{textcomp}
\usepackage{gensymb}
\usepackage{soul}
\usepackage{booktabs}
\usepackage{multirow}
\usepackage{arydshln}
\usepackage{graphicx}
\usepackage{subfig}
\usepackage{float}
\usepackage{caption}
\usepackage{tikz}
\usetikzlibrary{shapes}

\theoremstyle{definition}

\theoremstyle{remark}

\newcommand{\tmpframe}[1]{\fbox{#1}}

\biboptions{numbers,comma,round,square}
\graphicspath{ {./figs/} }

\journal{Elsevier}

\begin{document}
\begin{frontmatter}

\title{AI-Powered Automated Model Construction for Patient-Specific CFD Simulations of Aortic Flows}



\author[ndAME]{Pan Du\corref{coreq}}
\author[ndCSE]{Delin An\corref{coreq}}
\author[ndCSE]{Chaoli Wang}
\author[ndAME,cornell]{Jian-Xun Wang\corref{corxh}}

\address[ndAME]{Department of Aerospace and Mechanical Engineering, University of Notre Dame, Notre Dame, IN}
\address[ndCSE]{Department of Computer Science and Engineering, University of Notre Dame, Notre Dame, IN}
\address[cornell]{Sibley School of Mechanical and Aerospace Engineering, Cornell University, Ithaca, NY}
\cortext[coreq]{Co-first authors who contributed equally to this work.} 
\cortext[corxh]{Corresponding author.} 
\ead{jw2837@cornell.edu}

\begin{abstract}

Image-based modeling is essential for understanding cardiovascular hemodynamics and advancing the diagnosis and treatment of cardiovascular diseases. Constructing patient-specific vascular models remains labor-intensive, error-prone, and time-consuming, limiting their clinical applications. This study introduces a deep-learning framework that automates the creation of simulation-ready vascular models from medical images. The framework integrates a segmentation module for accurate voxel-based vessel delineation with a surface deformation module that performs anatomically consistent and unsupervised surface refinements guided by medical image data. By unifying voxel segmentation and surface deformation into a single cohesive pipeline, the framework addresses key limitations of existing methods, enhancing geometric accuracy and computational efficiency. Evaluated on publicly available datasets, the proposed approach demonstrates state-of-the-art performance in segmentation and mesh quality while significantly reducing manual effort and processing time. This work advances the scalability and reliability of image-based computational modeling, facilitating broader applications in clinical and research settings.

\end{abstract}

\end{frontmatter}


\clearpage

\section{Introduction}
\label{sec:intro}

Cardiovascular disease (CVD) remains one of the leading causes of mortality worldwide, accounting for millions of deaths annually, according to the World Health Organization (WHO)~\cite{csahin2022risk}. Effectively understanding and managing CVD requires advanced diagnostic tools capable of accurately characterizing complex hemodynamics within the cardiovascular system. While medical imaging modalities such as computed tomography (CT) and magnetic resonance imaging (MRI) provide high-resolution anatomical detail, they lack the capability to directly capture hemodynamics information (e.g., blood flow patterns, pressure, and wall shear stress fields) critical for understanding vascular function and pathology. To bridge this gap, image-based computational fluid dynamics (CFD) has emerged as a powerful computational paradigm that derives hemodynamic information from anatomical images via conservation laws. Although widely utilized in cardiovascular research, the clinical application of image-based CFD for diagnosis and surgical planning remains limited, largely due to the challenges associated with efficient and accurate model construction~\cite{steinman2002image,taylor2009patient,gray2018patient}.

Constructing patient-specific vascular models for image-based CFD involves multiple steps, including image segmentation, geometry modeling, and mesh generation for the computational domain, all of which are critical to ensuring the fidelity of the final simulation results. However, the standard workflow heavily relies on manual methods, making it highly labor-intensive and time-consuming. For instance, in SimVascular—a widely used cardiovascular simulation platform~\cite{updegrove2017simvascular}—experts delineate the centerline of blood vessels, segment the lumen boundary across individual 2D slices perpendicular to the centerline, and reconstruct the vascular geometry using B-splines. This process demands meticulous effort and often takes hours or even days to complete for a single patient, making it impractical for real-time or high-throughput applications~\cite{du2022deep,arzani2022machine}. Furthermore, the reliance on operator expertise introduces significant subjectivity, leading to variability between cases and potential errors such as inaccurate branch junctions and boundary delineation. These issues are further exacerbated by variations in image quality and acquisition protocols, which make it difficult to achieve reproducible and accurate geometries~\cite{du2022reducing}. Together, these limitations undermine the reliability of simulations and pose significant barriers to the integration of image-based modeling into clinical workflows.

Automating the process of constructing patient-specific vascular models is essential to overcome the limitations of manual workflows, particularly in terms of efficiency, consistency, and scalability. A fully automated pipeline can minimize operator dependency, reduce variability, and enable rapid generation of reproducible geometries, making image-based computational modeling more practical for clinical applications~\cite{ajam2017review}. Early attempts to automate this workflow primarily relied on traditional methods, such as deformable contour models (DCM)~\cite{rueckert1997automatic, das2006aortic,krissian2014semi,wang2017segmentation,ling2019fast,leventon2002statistical,he2008comparative}, active surface methods (ASM) \cite{jorstad2015neuromorph,mcinerney1995dynamic,terzopoulos1988constraints,terzopoulos1988symmetry,lareyre2019fully,bidhult2019new}, and level set methods~\cite{antiga2008image,osher2004level,volonghi2016automatic,kurugol2015automated,zhuge2006abdominal}. These techniques introduced some level of automation by optimizing geometric models to fit image data based on predefined energy functions. For example, DCMs minimized energy terms composed of internal forces promoting smoothness and external forces aligning the contour with image gradients~\cite{rueckert1997automatic,krissian2014semi}. ASMs extended DCMs to three-dimensional scenarios, enhancing their utility for complex structures. While effective in certain cases, these methods often struggled with complex vascular geometries and imaging noises. They also required precise initialization and parameter tuning, undermining their robustness and practicality in non-trivial scenarios~\cite{lareyre2019fully,bidhult2019new}. The level-set method offered another mathematical approach to segmentation by evolving a scalar field over time to extract zero-level isosurfaces that represent the boundaries of target vascular geometries~\cite{osher2004level,antiga2008image}. This evolution was guided by image features, such as gradients or region-based information, to delineate vascular structures more accurately. These methods achieved success in segmenting large vessels/ventricles and were incorporated into open-source platforms like the Vascular Modeling Toolkit~\cite{antiga2008image}. However, they also remained highly sensitive to initialization, imaging noise, and hyperparameters, often requiring extensive case-specific fine-tuning and user intervention. Notably, level-set methods tended to produce false detections, particularly in images with low peak signal-to-noise ratios (PSNR) or significant artifacts. Their reliance on threshold-based parameterization further limited their generalizability across diverse cases~\cite{kurugol2015automated, zhuge2006abdominal}.

The introduction of deep learning (DL) has significantly advanced medical image segmentation by offering data-driven solutions that outperform traditional methods in terms of accuracy and robustness~\cite{chen2020deep,jia2021learning}. DL-based segmentation models, particularly those employing convolutional neural networks (CNNs) and encoder-decoder architectures, leverage large annotated datasets to directly learn the mapping between image inputs and semantic segmentation outputs. These methods have demonstrated remarkable success in segmenting cardiovascular structures, including ventricles, arteries, and capillaries~\cite{wolterink2016dilated,bai2018recurrent,xia2019automatic,vigneault2018omega,wolterink2018automatic,baumgartner2018exploration}. However, a significant limitation of most DL-based segmentation approaches is their focus on identifying the region of interest using binary voxel classifications, without facilitating the creation of 3D computational meshes that can be used for downstream CFD or fluid-structure interaction (FSI) simulations. In order to use the voxel-based outputs for computational modeling, additional post-processing steps, such as marching cubes~\cite{newman2006survey}, are required to generate smooth surfaces. This process can introduce artifacts, reduce geometric fidelity, and undermine the accuracy of simulations. To address these challenges, recent studies have explored integrated frameworks that combine voxel segmentation and surface reconstruction. For example, deep active surface models (DASM) employ graph convolutional neural networks (GNNs) to iteratively deform surfaces by simultaneously minimizing energy losses and surface regularization terms~\cite{zhao2022segmentation,wickramasinghe2021deep}. Similarly, methods like Voxel2Mesh integrate voxel-based segmentation with surface deformation, using encoder-decoder networks for voxel predictions and GNNs to represent and deform surface mesh~\cite{kong2021deep,bongratz2022vox2cortex,wickramasinghe2020voxel2mesh}. In these methods, features extracted from the encoder are projected onto the surface vertices and integrated into the hidden layers of the GNN to guide deformations. Despite their promise, these approaches face challenges when dealing with multi-branched vascular geometries, as they require handling significant non-rigid transformations. Interestingly, non-rigid deformation has been extensively studied in the computer vision community, where various DL-based shape registration algorithms have shown impressive performance in aligning complex geometries~\cite{deng2022survey,eisenberger2021neuromorph,amor2022resnet,eisenberger2020smooth}. Some of these methods incorporate physics-based deformation models, such as large deformation diffeomorphic metric mapping (LDDMM)~\cite{beg2005computing}, to ensure smooth transformations. For instance, Amor et al.\ \cite{amor2022resnet} combined a ResNet structure with LDDMM to align geometries from datasets that included human organs such as the heart and ventricles. However, despite their success in broader shape registration problems, these methods have yet to see extensive application in surface reconstruction for cardiac image segmentation.

Another critical limitation of most DL-based segmentation models is their reliance on manually crafted surface labels, which are often assumed to be the gold standard despite inherent imperfections. For instance, in SimVascular, vascular geometries are reconstructed using non-uniform rational B-splines (NURBS) fitted to discrete slices along manually identified vessel centerlines~\cite{updegrove2017simvascular}. While this approach provides an approximation of the vascular surface, the majority of the surface points do not directly correspond to image-derived features, leading to inaccuracies. Additionally, SimVascular employs blending operations to smooth junctions between vessels~\cite{updegrove2016boolean}, introducing artifacts that might be inconsistent with the image gradients. These imperfections in training labels propagate through segmentation pipelines, limiting the accuracy and generalizability of supervised DL models.

Despite these challenges, integrating DL-based segmentation with robust surface reconstruction frameworks provides a promising pathway to enhance the automation and accuracy of image-based cardiovascular modeling. In this work, we propose a novel DL framework that unifies voxel segmentation and surface deformation into a single, cohesive pipeline for the automated construction of patient-specific models for image-based CFD simulations of aortic flows. The framework begins with a Laplacian-of-Gaussian-based Bayesian network (LoGB-Net), a supervised segmentation module designed to enhance robustness and segmentation accuracy across vascular structures of varying scales. By incorporating uncertainty quantification (UQ), the LoGB-Net improves the delineation of subtle or blurred vessel boundaries commonly encountered in medical imaging. The voxel-based segmentation output is then refined using a GNN-LDDMM surface deformation module, ensuring smooth and anatomically consistent transformations. Guided by unsupervised learning objectives, including alignment and image-gradient-driven energy terms, the surface deformation model transforms voxel-based outputs into anatomically realistic 3D surfaces suitable for CFD simulations. Together, these components address key limitations of existing methods, offering a robust and automated solution for generating simulation-ready vascular models. This unified framework significantly improves efficiency, consistency, and accuracy, advancing the potential for image-based cardiovascular analysis in both clinical and research applications.


\section{Results}
\label{sec:result}

\subsection{Overview of proposed framework and results}

In this work, we present a novel integrated DL framework for the automated construction of patient-specific models for image-based CFD simulations of aortic flows. The framework unifies voxel segmentation and surface deformation into a single cohesive pipeline, significantly enhancing both the accuracy and efficiency of model construction. The process begins with a voxel-based segmentation module leveraging a novel machine learning model called the LoGB-Net. This supervised module is specifically designed to improve segmentation accuracy across vascular structures of varying scales. By incorporating Bayesian principles, the LoGB-Net offers robustness to noise and uncertainty quantification (UQ), which is particularly valuable for identifying subtle or blurred vessel boundaries frequently encountered in medical imaging. The output of this module is a high-resolution binary voxel map, which serves as the foundation for surface reconstruction. 

To transform the voxel-based segmentation into anatomically realistic surface models, we developed a surface deformation model where the LDDMM algorithm is integrated with GNNs, enabling the framework to handle complex non-rigid transformations while maintaining smoothness and fidelity to the underlying image structures. A key innovation is the elimination of traditional surface regularization terms, as the LDDMM inherently enforces smooth and physically plausible deformations. The surface deformation process is guided by unsupervised learning objectives, including an energy term that aligns the surface with image gradients to ensure geometric consistency and anatomical accuracy, along with additional terms that prevent unrealistic regional deformation.  

The proposed framework combines supervised and unsupervised learning paradigms. The LoGB-Net module operates under a supervised setting, requiring labeled voxel data for training. These training labels are exclusively used for voxel segmentation, enabling the model to adapt to diverse imaging conditions and accurately delineate vascular regions. In contrast, the surface deformation module operates under an unsupervised paradigm, relying on alignment loss and image-gradient energy terms to guide the reconstruction process without manually annotated training labels. Notably, manually constructed models, generated using SimVascular, are used only for comparison and benchmarking purposes. Details on the training and testing datasets are provided in Section~\ref{sec:data}. 

\begin{figure}[tp!]
    \centering
    \tmpframe{\includegraphics[width=\textwidth]{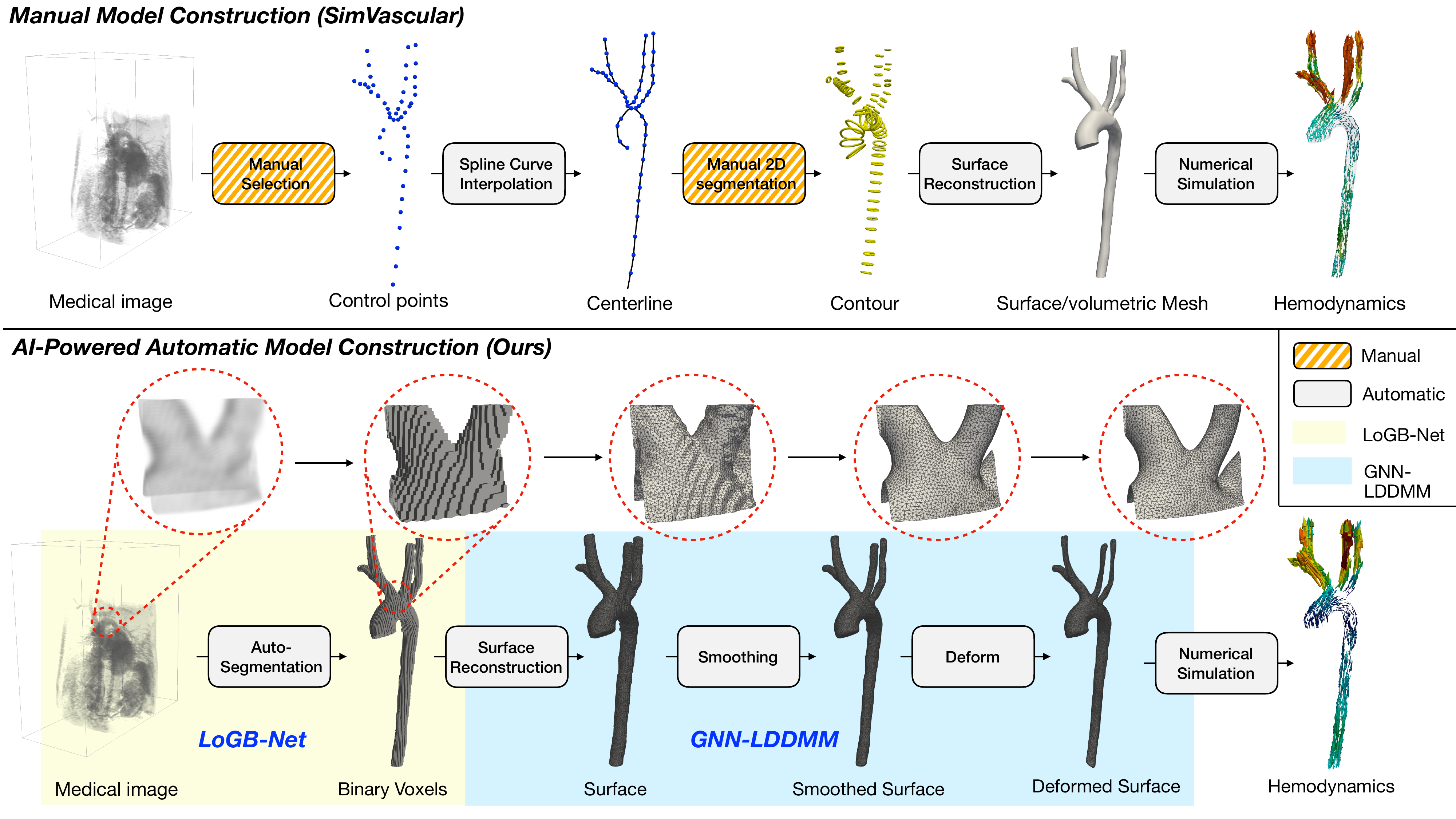}}
    \caption{Comparison of manual and AI-powered automatic model construction workflows for image-based simulations. Representative results at each step of our proposed framework are shown in the bottom panel.}
    \label{fig:res-overview}
\end{figure}
Figure~\ref{fig:res-overview} provides a schematic overview of the proposed framework applied to a representative testing sample, compared with the traditional manual model construction workflow in SimVascular~\cite{updegrove2017simvascular}. The top panel of Figure~\ref{fig:res-overview} outlines the manual workflow, which includes manually selecting vessel centerlines, interpolating using splines, and manually segmenting 2D slices to reconstruct the vascular surface. The reconstructed surface is converted into a volumetric mesh suitable for CFD simulations. Although this manual process can yield high-quality results, it is labor-intensive, time-consuming, and prone to operator variability.

The bottom panel of Figure~\ref{fig:res-overview} showcases the automated results of the proposed framework. The LoGB-Net module produces a binary voxel segmentation of the vascular lumen that serves as a coarse, voxelized representation of the vascular structure. Although accurate in capturing the overall shape, this voxel-based output displays blocky discrete features inherent to its nature. The GNN-LDDMM module then refines this preliminary surface, smoothing and deforming it into an anatomically realistic representation. The final reconstructed surface closely matches the underlying anatomical structures and is ready for CFD simulations. Although the automated framework demonstrates strong agreement with the manual results, minor discrepancies are evident in fine geometric details. These differences, shown in Figure~\ref{fig:res-overview}, will be analyzed in subsequent sections according to the medical image input. Notably, the proposed framework achieves a substantial reduction in processing time while ensuring anatomical consistency and reproducibility. This balance of efficiency and accuracy marks a significant advance in patient-specific cardiovascular modeling.

\subsection{Voxel segmentation evaluation}

The binary voxel segmentation performance of the proposed LoGB-Net model was evaluated against ten state-of-the-art (SOTA) DL-based segmentation methods, including FPN~\cite{Lin_2017_CVPR}, U-Net 3D~\cite{cciccek20163d}, PSPNet~\cite{Zhao2017}, nnUNet~\cite{isensee2021nnu}, Attention-UNet~\cite{oktay2018attention}, MISSFormer~\cite{huang2021missformer}, Swin-UNET~\cite{Cao2023}, TransUNet~\cite{Chen2021}, UNETR~\cite{Hatamizadeh2022}, and UNETR++~\cite{shaker2023unetr}. The proposed LoGB-Net features a ``shape stream'' module specifically designed to detect vessels of varying diameters, which is critical for accurately modeling multi-branch aortic geometries as well as other tubular cardiovascular structures, such as cerebral and pulmonary arteries. More details can be found in Section~\ref{sec:meth: LoGB}. Among the baseline methods, UNETR employs residual blocks to improve feature extraction, while Attention-UNet utilizes attention mechanisms to focus on critical regions selectively. These SOTA methods, along with the others evaluated, represent the current advances in 3D medical image segmentation and provide a robust benchmark for assessing LoGB-Net's performance.

\begin{figure}[t!]
    \centering
    \includegraphics[width=1.0\textwidth]{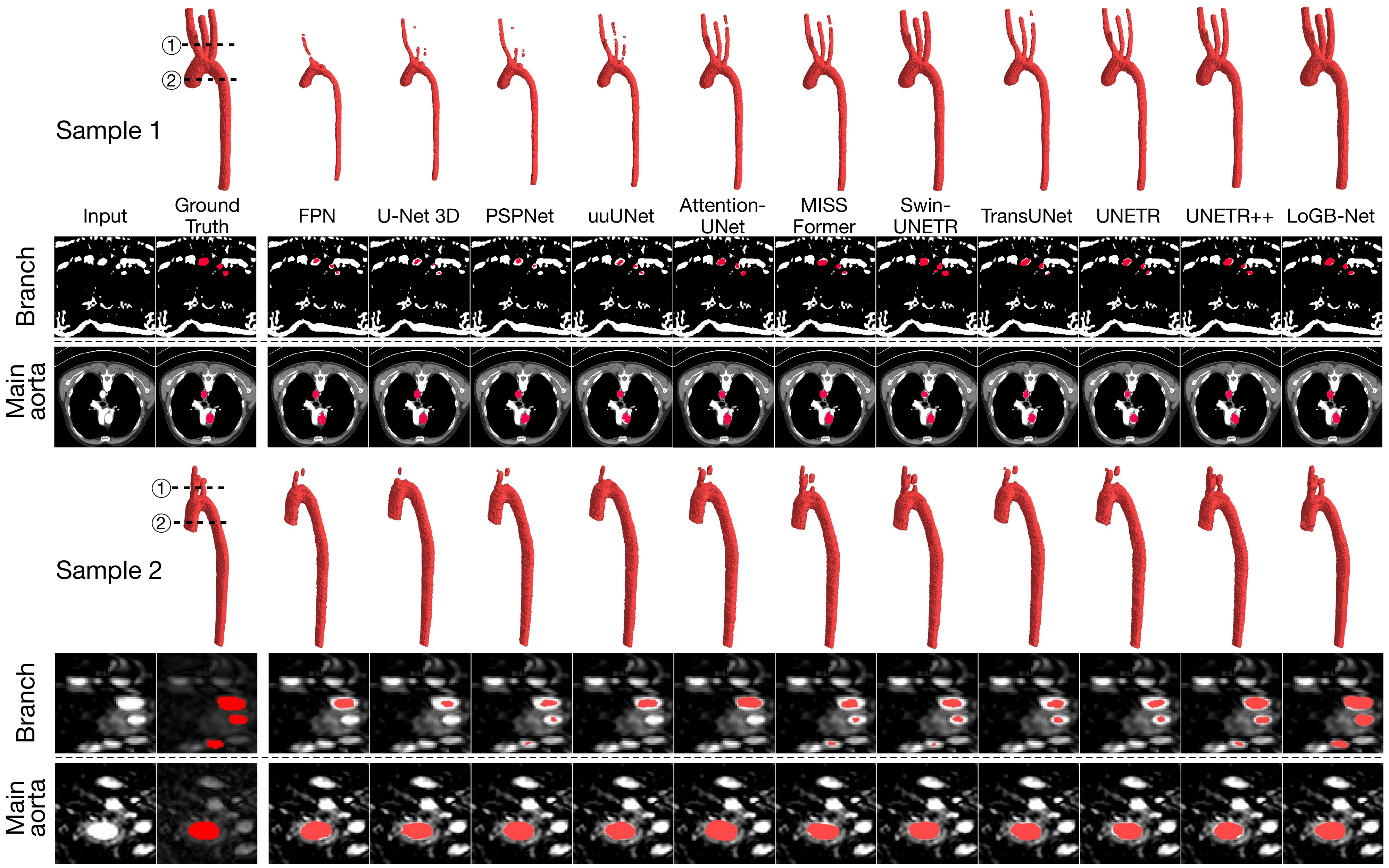}
    \caption{Comparison of segmentation results for two representative samples across different models. The top row for each sample shows 3D reconstructions of segmented aortic geometries, while the lower rows provide cross-sectional views at the supra-aortic branches (Branch) and the main aorta (Main aorta). LoGB-Net demonstrates superior performance, producing anatomically accurate and continuous segmentations, particularly in challenging regions such as the supra-aortic branches.}
    \label{fig:segmentation}
\end{figure}
Figure~\ref{fig:segmentation} compares segmentation results across models for two representative samples. For both samples, the top row shows 3D reconstructions of segmented vascular structures, while the bottom rows provide cross-sectional views at two critical locations: the supra-aortic branches (Branch) and the main aorta (Main aorta). While most models perform well in segmenting the main aorta, significant variability is observed in segmenting smaller, more complex branches. Models like FPN, U-Net 3D, PSPNet, and nnUNet frequently produce incomplete or disconnected segmentations for the upper branches, such as the right and left common carotid arteries and the subclavian arteries. Attention-based models, including Attention-UNet, MISSFormer, Swin-UNETR, and TransUNet, improve continuity but still fail to consistently achieve anatomically accurate results, particularly for challenging regions in test sample 2. By contrast, UNETR++ and our LoGB-Net demonstrate superior performance, with LoGB-Net consistently producing the most accurate and continuous segmentations. The inclusion of the LoGB stream module enhances its ability to detect vessels across a wide range of diameters, contributing to its improved performance in segmenting smaller and more complex structures.

To further assess the performance quantitatively, three established metrics, Dice Similarity Coefficient (Dice), Average Surface Distance (ASD), and Hausdorff Distance, are computed. The Dice coefficient quantifies the overlap between predicted segmentation and the ground truth, with values closer to one indicating better agreement. ASD calculates the mean distance between corresponding surface points of the predicted and actual boundaries, while Hausdorff distance measures the maximum distance, emphasizing outlier errors. These metrics provide a comprehensive evaluation of segmentation accuracy and geometric fidelity. Mathematical details of these metrics are provided in Section~\ref{sec:train and test}. Table~\ref{tab:quantitative1} provides quantitative results for all testing samples, reinforcing the visual findings. LoGB-Net achieves the highest Dice coefficients (92.7\% for branch vessels and 93.7\% for the main aorta), the lowest ASD values (0.678 mm for branch vessels and 0.682 mm for the main aorta), and competitive Hausdorff distances for both vessel types. UNETR++ is the closest competitor but still lags behind LoGB-Net across all metrics, particularly for smaller branches. 
\begin{table*}[htb]
    \centering
    \fontsize{30}{20}\selectfont
    \resizebox{\textwidth}{!}{
    \begin{tabular}{ccccccccccccc}
    \toprule
    \multirow{2}{*}{} & \multicolumn{1}{c}{Type} & \multicolumn{4}{c}{CNN-based} & \multicolumn{6}{c}{Attention-based} & \multicolumn{1}{c}{Ours}\\
    \cmidrule(lr){3-6} \cmidrule(lr){7-12} \cmidrule(lr){13-13}
    & Metric & FPN & U-Net 3D & PSPNet & nnUNet & Attention-UNet & MISSFormer & Swin-UNETR & TransUNet & UNETR & UNETR++ & LoGB-Net\\
    \midrule
    \multirow{3}*{\rotatebox{90}{Branch}} & Dice $\uparrow$ & 0.72$\pm$0.02 & 0.76$\pm$0.04 & 0.78$\pm$0.04 & 0.77$\pm$0.07 & 0.82$\pm$0.03 & 0.88$\pm$0.02 & 0.90$\pm$0.02 & 0.82$\pm$0.02 & 0.88$\pm$0.03 & 0.89$\pm$0.04 & \textbf{0.93}$\pm$\textbf{0.01} \\
        & ASD $\downarrow$ & 1.52$\pm$0.32 & 1.38$\pm$0.31 & 1.33$\pm$0.30 & 1.21$\pm$0.32 & 0.93$\pm$0.21 & 0.81$\pm$0.20 & 0.79$\pm$0.25 & 0.93$\pm$0.28 & 0.70$\pm$0.17 & 0.68$\pm$0.16 & \textbf{0.68}$\pm$\textbf{0.17} \\
        & Hausdorff $\downarrow$ & 8.49$\pm$3.25 & 8.32$\pm$4.02 & 9.77$\pm$3.77 & 8.94$\pm$2.55 & 6.73$\pm$0.29 & 6.90$\pm$0.08 & 6.32$\pm$0.13 & 6.45$\pm$0.24 & 6.14$\pm$0.70 & \textbf{6.03}$\pm$\textbf{0.36} & 6.23$\pm$0.96 \\
    \hdashline
    \multirow{3}*{\rotatebox{90}{Main}} & Dice $\uparrow$ & 0.73$\pm$0.02 & 0.77$\pm$0.03 & 0.78$\pm$0.03 & 0.78$\pm$0.03 & 0.85$\pm$0.02 & 0.90$\pm$0.02 & 0.91$\pm$0.03 & 0.90$\pm$0.02 & 0.90$\pm$0.04 & 0.91$\pm$0.02 & \textbf{0.94}$\pm$\textbf{0.01} \\
    & ASD $\downarrow$ & 1.45$\pm$0.36 & 1.34$\pm$0.34 & 1.34$\pm$0.31 & 1.40$\pm$0.31 & 0.88$\pm$0.23 & 0.92$\pm$0.23 & 0.80$\pm$0.22 & 0.93$\pm$0.31 & 0.69$\pm$0.20 & 0.74$\pm$0.19 & \textbf{0.68}$\pm$\textbf{0.20} \\
    & Hausdorff $\downarrow$ & 10.25$\pm$3.78 & 11.56$\pm$3.93 & 9.99$\pm$4.20 & 9.29$\pm$2.33 & 6.94$\pm$0.31 & \textbf{6.13}$\pm$\textbf{0.19} & 7.02$\pm$0.07 & 6.88$\pm$0.25 & 6.42$\pm$0.45 & 6.32$\pm$0.24 & 6.32$\pm$1.49 \\
    \bottomrule
    \end{tabular}}
    \caption{Quantitative results of different methods. The best results are highlighted in bold. The results are reported as mean $\pm$ standard deviation. The experiment was conducted five times with varying random seeds to evaluate the outcomes.}
    \label{tab:quantitative1}
\end{table*}

In summary, the qualitative and quantitative analyses demonstrate that LoGB-Net outperforms existing SOTA methods, showing great capability to reliably segment complex, multi-branch vascular geometries with minimal manual intervention.


\subsection{Surface reconstruction analysis}

Following the voxel-based segmentation generated by LoGB-Net, an essential next step is to transform the segmented voxelized geometry into simulation-ready meshes for downstream CFD analysis. To address this need, we developed a surface reconstruction pipeline using unsupervised learning. Unlike traditional methods, such as those employed in SimVascular, which are time-consuming and prone to operator variability, our automated pipeline efficiently generates simulation-ready models with minimum manual intervention, greatly enhancing scalability and reproducibility. Beyond significantly reducing labor time, our approach achieves geometric accuracy that often surpasses manual methods. This is attributed to our unique surface deformation algorithm, which fine-tunes the reconstructed surface by iteratively aligning it with the underlying image gradients. Unlike manual workflows that interpolate between a limited number of annotated cross-sections, our algorithm dynamically adjusts every surface point during an unsupervised optimization process, ensuring comprehensive alignment with the entire medical images. This eliminates common inaccuracies in manual methods, such as oversmoothing caused by NURBS interpolation or artifacts at multi-branch intersections due to boolean operations. Here, we present a detailed comparison of the resulting surfaces and corresponding CFD solutions between our deformed surface and manual ones derived from open-source, expert-verified datasets.

\begin{figure}[t!]
    \centering
    \includegraphics[width=1.0\textwidth,trim=9 15 9 9,clip]{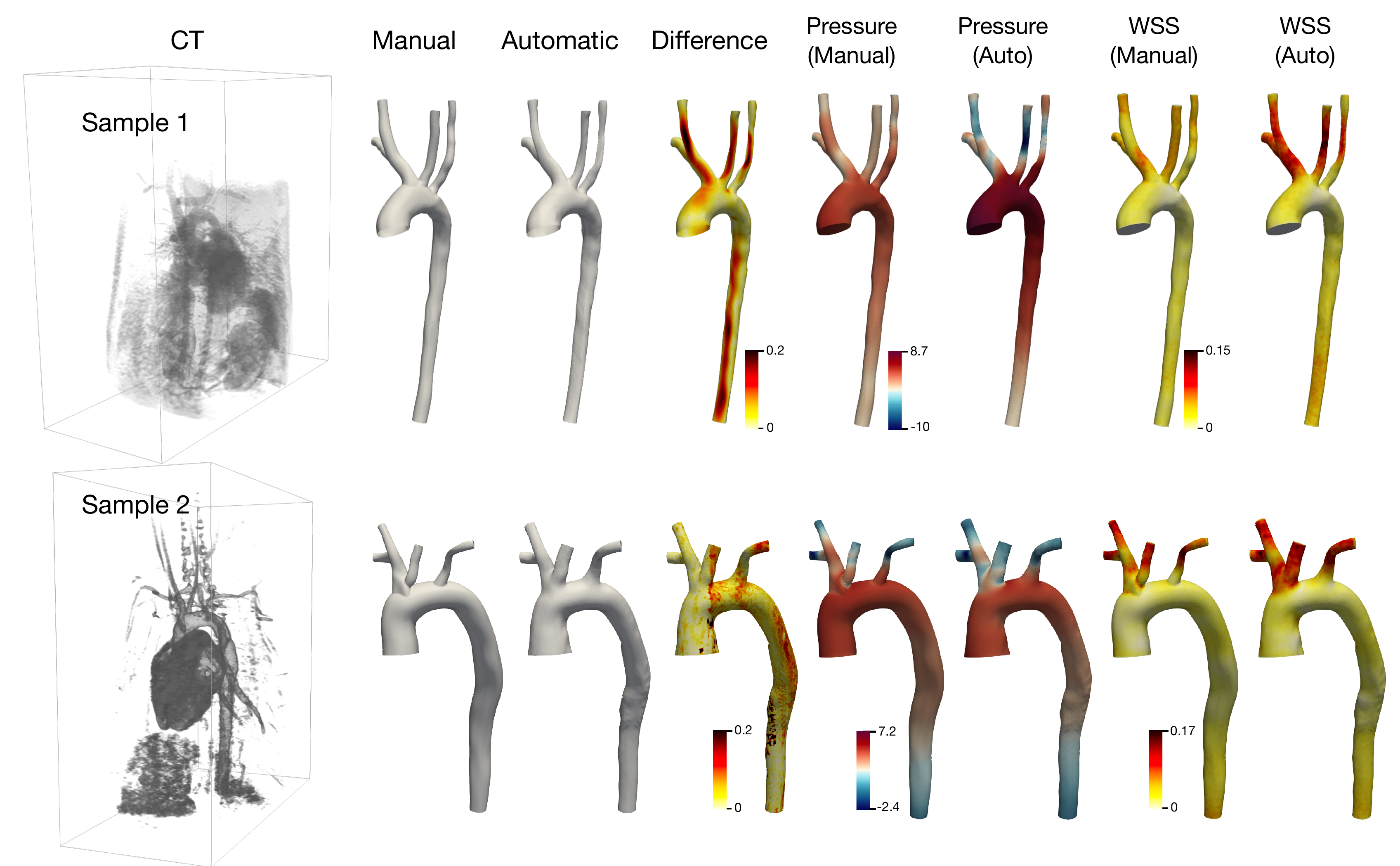}
    \caption{Comparison of manual and automated surface reconstruction for two representative samples. Columns display the input CT image, manual and automated reconstructed surfaces, geometric difference maps, and the resulting pressure and wall shear stress (WSS) distributions from CFD simulations.}
    \label{fig:deform1}
\end{figure}
As shown in Figure~\ref{fig:deform1}, both manual and automated methods successfully reconstruct vascular geometries, capturing the essential regions within the CT dataset. On visual inspection, the reconstructed surfaces appear highly similar, with the automated method matching the global geometric features produced by manual reconstruction. A detailed difference map, shown as the distance contour, highlights some subtle geometric discrepancies, which are predominantly localized in branch regions and areas with more complex vessel intersections. These differences are attributed to the inherent smoothing of manual methods and the comprehensive point-wise optimization in our approach, which ensures tighter adherence to image features. Despite minor geometric differences, the impact on CFD results is substantial, as evidenced in the pressure and wall shear stress (WSS) distributions (Figure~\ref{fig:deform1}). The automated pipeline captures finer variations in both pressure and WSS patterns, yielding higher maximum values and more nuanced distributions along the streamwise direction. These hemodynamic factors, which are critical biomarkers for cardiovascular diagnostics, highlight the importance of precise geometry reconstruction and uncertainty quantification.

\begin{figure}[t!]
    \centering
    \includegraphics[width=1.0\textwidth,trim=9 15 9 9,clip]{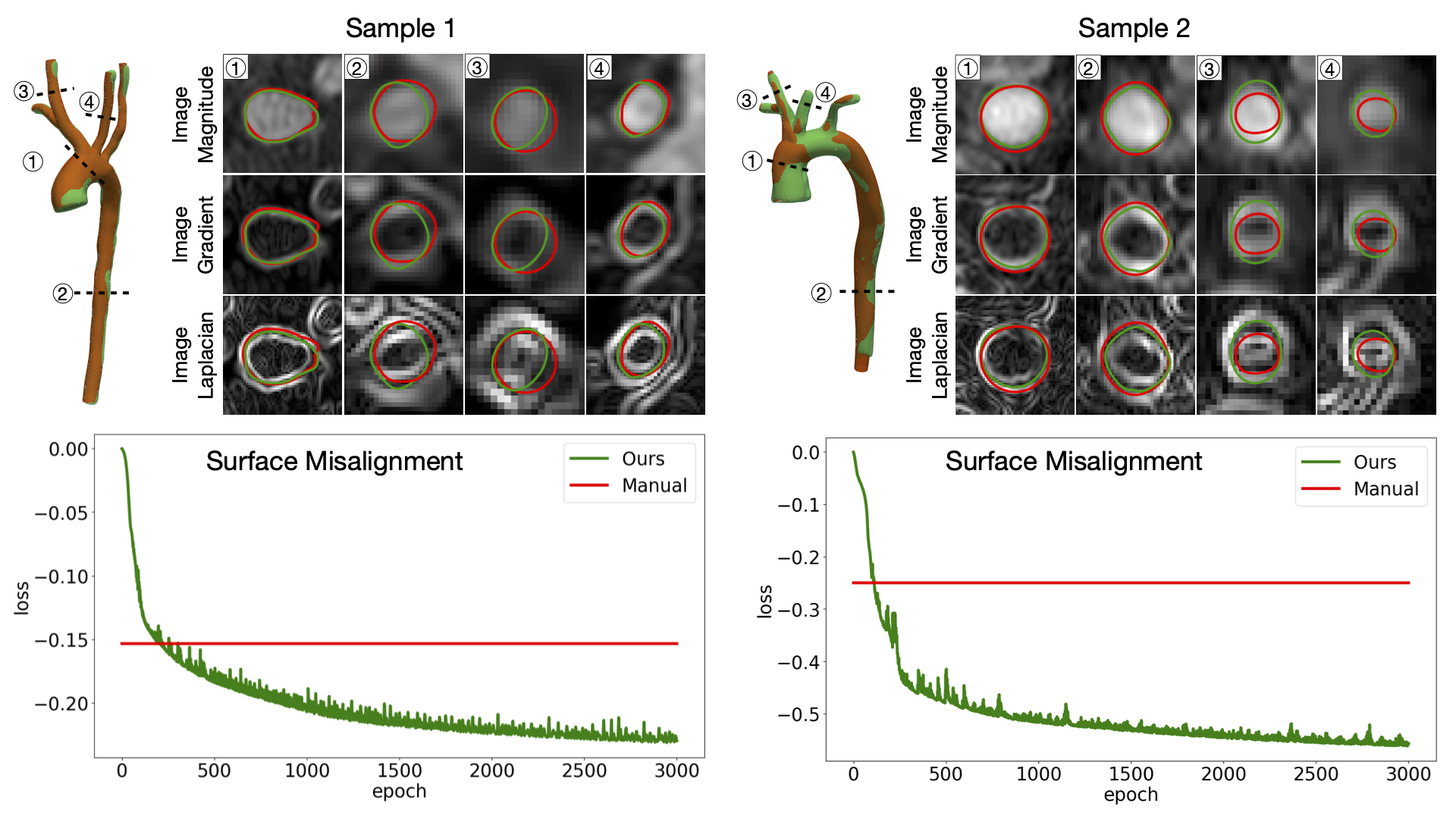}
    \caption{Detailed cross-sectional comparison of reconstructed surfaces and source images for two representative samples. For each cross-section, overlays of the manual surface (red) and the automated surface (green) are shown against three background references: image magnitude, image gradient, and image Laplacian. The lower panels present the surface misalignment loss during training, where the automated surfaces achieve rapid convergence to lower loss values compared to static, manual surfaces.}
    \label{fig:deform2}
\end{figure}
Figure~\ref{fig:deform2} provides a detailed cross-sectional comparison between manual and automated surfaces at four representative locations (labeled 1 through 4) for two samples. For each cross-section, overlays of the surfaces are shown against three background images: image magnitude, image gradient, and image Laplacian. In the magnitude images, the automated surface (green) consistently aligns slightly better with the vessel boundaries, while the manual surface (red) often deviates due to oversmoothing. For example, in Section 1 of Sample 1, the manual surface fails to fully capture the narrow region of the vessel wall, leading to an underrepresentation of the actual geometry. Similarly, in Section 4 of Sample 2, the manual surface erroneously extends into adjacent regions, an artifact caused by interpolation errors in the manual workflow. The gradient images emphasize regions of high-intensity change, which correspond to vessel boundaries. Here, the automated surface matches closely to the middle of the high-gradient regions, as seen in Sections 2 and 3 of both samples. By contrast, the manual surface slightly deviates, particularly in high-curvature or branching areas, such as Section 3 of Sample 2, where it cuts into the vessel interior. Laplacian images, which highlight transitions between regions of different intensities, can further show the alignment between reconstructed surfaces and images. 

The lower panels in Figure~\ref{fig:deform2} quantitatively evaluate surface alignment through the energy loss term, which measures misalignment between the reconstructed surface and the image gradients during the deformation process. For both samples, the automated pipeline rapidly reduces the loss over the course of training and converges to significantly lower values compared to the manually reconstructed surfaces. Notably, the manual surfaces exhibit a static loss value, reflecting their inability to adjust beyond the initial annotations. In contrast, the automated surfaces continuously optimize their alignment throughout the unsupervised optimization process, achieving superior adherence to image features by approximately the $200$th epoch. 
These results demonstrate that our automated pipeline not only automates surface reconstruction, significantly reducing the labor-intensive nature of manual workflows, but also delivers geometries with enhanced fidelity to the source images.

\subsection{Uncertainty estimation and propagation}

\begin{figure}[htp!]
    \centering
    \includegraphics[width=0.92\textwidth,trim=9 15 9 9,clip]{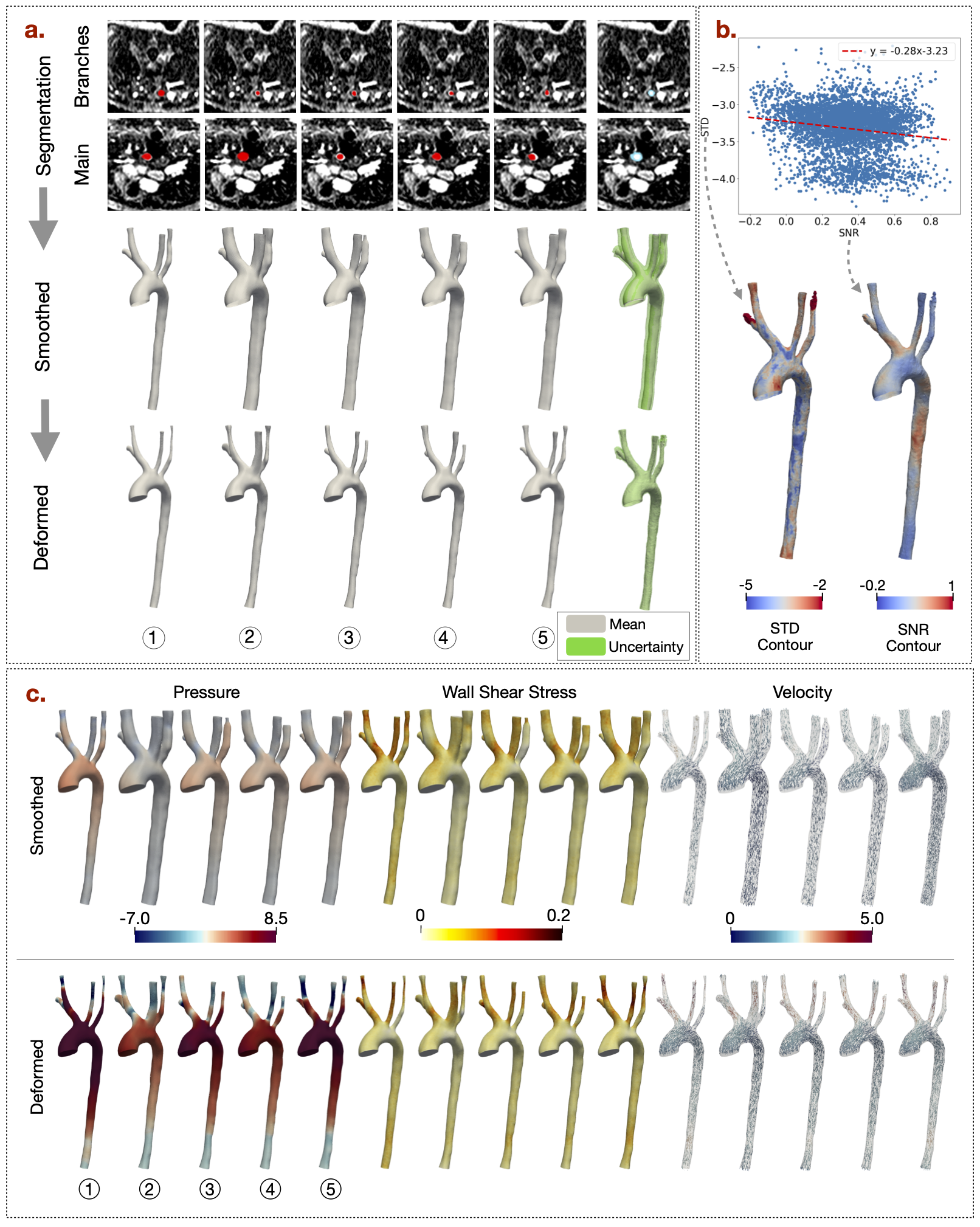}
    \caption{Uncertainty estimation and propagation through surface reconstruction and CFD simulation. (a) Segmentation uncertainty visualized across voxel-based predictions, smoothed geometries, and deformed surfaces, with uncertainty maps highlighting regions of variability. (b) Correlation between uncertainty (STD) and image quality (SNR), with corresponding surface contours. (c) Comparison of hemodynamic quantities from CFD simulations for smoothed and deformed surfaces.)}
    \label{fig:uq}
\end{figure}

This section explores the uncertainty quantification (UQ) capabilities of the LoGB-Net model and demonstrates how uncertainties propagate through the GNN-LDDMM surface deformation and CFD simulation steps. Figure~\ref{fig:uq}a showcases random segmentation realizations obtained by sampling the learned probabilistic distributions of the model parameters of the LoGB-Net. The top row illustrates five representative voxel-based segmentation results for both the main aorta and the branches, with the rightmost column depicting uncertainty regions that highlight areas of potential edge variability. These uncertainty estimates are crucial for understanding the confidence levels of the segmentation outputs. 

The second row presents the smoothed geometries derived directly from the voxel-based outputs using a standard smoothing algorithm to produce simulation-read meshes. The uncertainties from the segmentation step persist, particularly near complex vessel regions (see green bands in the rightmost column). The bottom row demonstrates the results after applying the GNN-LDDMM deformation module, which aligns the surface geometry with underlying medical image gradients. Compared to the smoothed geometries, the deformed results exhibit significantly reduced uncertainty, particularly in the main aorta, as seen in the tighter green bands around the mean surface. 

Figure~\ref{fig:uq}b analyzes the relationship between segmentation uncertainty and image quality. The scatter plot correlates the log of the standard deviation (STD) of uncertainty with the signal-to-noise ratio (SNR) across all points on the mean surface. A negative correlation is observed, with a regression slope of -0.28, indicating that regions with higher SNR exhibit lower uncertainty. This trend is further supported by the STD and SNR contour plots overlaid on the mean surface. Regions with high uncertainty, such as the distal ends of branch vessels, correspond to areas of lower SNR, highlighting the impact of poor image quality on segmentation reliability.

Figure~\ref{fig:uq}c compares CFD solutions for pressure, wall shear stress (WSS), and velocity fields across the smoothed and deformed surface ensembles. The results reveal substantial differences in hemodynamic outcomes, even for minor geometric discrepancies. Deformed surfaces yield higher average pressure and WSS values compared to smoothed ones, reflecting the improved anatomical alignment achieved by the deformation module. While the main aorta geometries converge closely after deformation, branch geometries still exhibit notable variations due to persistent noise in these regions. This variability highlights a limitation of the current deformation model in handling severe noise, which is often encountered in real-world imaging datasets.

Overall, our framework leverages UQ to generate ensembles of potential segmentation outcomes, moving beyond single-point predictions and offering a probabilistic understanding of geometric variability. The GNN-LDDMM deformation module plays a crucial role in refining these surfaces, reducing uncertainty, and improving alignment with medical images.

\subsection{Framework ablation study}

\subsubsection{Ablation study of the segmentation module}

To evaluate the necessity and effectiveness of the key components in the LoGB-Net segmentation model, we conducted a comprehensive ablation study focusing on the LoG stream, Bayesian framework, and balanced gate (see Table~\ref{tab:quantitative2}). The segmentation performance was assessed under various configurations, with each component removed individually. The results in the left three columns demonstrate significant performance degradation when any of these components is excluded. This finding demonstrates the unique contribution of each subcomponent in the overall architecture. The LoG stream, serving as the core feature of LoGB-Net, has the most pronounced impact on performance. Its multiscale kernel setup enhances the model’s ability to detect vessels of varying diameters, making it indispensable for handling complex vascular geometries. The balanced gate addresses foreground-background imbalance during training, ensuring stable and robust optimization. Meanwhile, the Bayesian framework improves robustness by learning the probability distribution of kernel parameters, which captures the inherent variability in medical image data.

To further understand the role of the LoG stream, we experimented with varying the number of LoG layers, represented as $\mathcal{L}(1)$ to $\mathcal{L}(5)$ in Table~\ref{tab:quantitative2}. The results show a clear trend: as the number of layers increases, performance improves across all metrics. This demonstrates that deeper hierarchical structures provide more effective multiscale vessel detection. However, performance appears to plateau at $\mathcal{L}(5)$, suggesting that additional layers may yield diminishing returns in this context.

\begin{table*}[t!]
  \caption{Quantitative results of the LoGB-Net's ablation study. $\mathcal{L}_{LoG}^-$, $\mathcal{L}_{Bay}^-$, and $\mathcal{L}_{Gate}^-$ represent LoGB-Net without LoG module, Bayesian optimization, and balanced gate, respectively. $\mathcal{L}(1)$, $\mathcal{L}(2)$, $\mathcal{L}(3)$, $\mathcal{L}(4)$, and $\mathcal{L}(5)$ represent LoGB-Net with 1, 2, 3, 4, and 5 LoG layers, respectively.}
  \centering
  \resizebox{\textwidth}{!}{
      \begin{tabular}{cccccccccc}
          \toprule
                                               & Metric                 & $\mathcal{L}_{LoG}^-$ ($\mathcal{L}(0)$) & $\mathcal{L}_{Bay}^-$ & $\mathcal{L}_{Gate}^-$ & $\mathcal{L}(1)$ & $\mathcal{L}(2)$ & $\mathcal{L}(3)$ & $\mathcal{L}(4)$ & $\mathcal{L}(5)$ (Ours)           \\
          \midrule
          \multirow{3}*{\rotatebox{90}{SA}} & Dice $\uparrow$        & 0.735$\pm$0.033                          & 0.863$\pm$0.031       & 0.896$\pm$0.012        & 0.852$\pm$0.033  & 0.863$\pm$0.018  & 0.872$\pm$0.011  & 0.893$\pm$0.033  & \textbf{0.927}$\pm$\textbf{0.011} \\
                                               & ASD $\downarrow$       & 1.384$\pm$0.334                          & 0.810$\pm$0.214       & 0.801$\pm$0.179        & 1.291$\pm$0.123  & 1.226$\pm$0.202  & 0.922$\pm$0.244  & 0.772$\pm$0.202  & \textbf{0.678}$\pm$\textbf{0.168} \\
                                               & Hausdorff $\downarrow$ & 8.271$\pm$2.276                          & 7.993$\pm$2.112       & 7.909$\pm$1.093        & 8.203$\pm$1.903  & 8.014$\pm$1.221  & 7.027$\pm$0.721  & 6.626$\pm$0.121  & \textbf{6.225}$\pm$\textbf{0.957} \\
          \hdashline
          \multirow{3}*{\rotatebox{90}{MA}} & Dice $\uparrow$        & 0.741$\pm$0.022                          & 0.871$\pm$0.033       & 0.882$\pm$0.021        & 0.866$\pm$0.014  & 0.865$\pm$0.009  & 0.877$\pm$0.036  & 0.923$\pm$0.003  & \textbf{0.937}$\pm$\textbf{0.006} \\
                                               & ASD $\downarrow$       & 1.399$\pm$0.123                          & 0.961$\pm$0.120       & 0.933$\pm$0.200        & 1.105$\pm$0.239  & 0.979$\pm$0.103  & 0.933$\pm$0.210  & 0.727$\pm$0.121  & \textbf{0.682}$\pm$\textbf{0.197} \\
                                               & Hausdorff $\downarrow$ & 8.335$\pm$2.996                          & 8.220$\pm$2.003       & 8.011$\pm$1.898        & 8.277$\pm$2.516  & 8.100$\pm$2.274  & 7.965$\pm$1.001  & 7.482$\pm$1.112  & \textbf{6.322}$\pm$\textbf{1.485} \\
          \bottomrule
      \end{tabular}}
  \label{tab:quantitative2}
\end{table*}

\subsubsection{Ablation study of the GNN-LDDMM deformation module}

\begin{figure}[t!]
    \centering
    \includegraphics[width=0.99\textwidth,trim=9 15 9 9,clip]{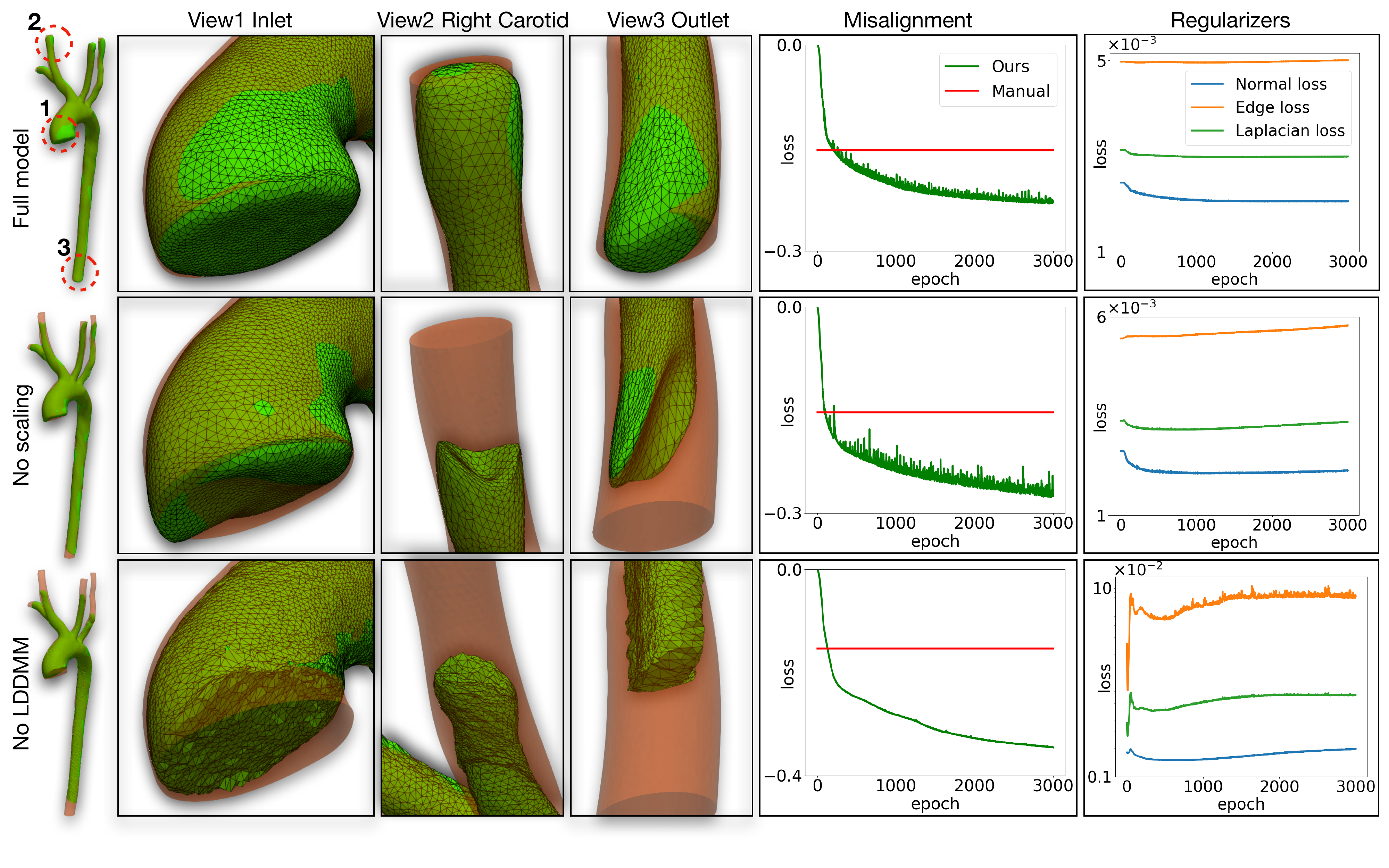}
    \caption{Ablation study for deformation model: mesh comparison in different views (first to fourth columns), Loss curve (fifth column)}
    \label{fig:ablation_deform}
\end{figure}

To evaluate the necessity and effectiveness of key components within the GNN-LDDMM deformation model, we conducted an ablation study, as illustrated in Figure~\ref{fig:ablation_deform}. Specifically, we examined the impact of removing critical elements such as the LDDMM module and the scaling component. These investigations provide insight into the importance of these elements in ensuring geometrically accurate and physically plausible surface deformations for subsequent CFD analysis.
The left four columns of Figure~\ref{fig:ablation_deform} display surface visualizations under different ablation scenarios. The absence of the scaling component leads to retraction of inlets and outlets, causing unrealistic geometries. This occurs because image gradients alone do not sufficiently constrain surface points in these areas, allowing them to drift toward regions with locally lower misalignment energy. The scaling component addresses this by stabilizing these regions, preserving vessel lengths and anatomical consistency. Furthermore, removing the LDDMM module results in substantial surface faults, particularly at the inlet and outlet regions. This deficiency arises because LDDMM acts as a strong prior for maintaining surface continuity and anatomical realism, which is crucial for ensuring the feasibility of downstream CFD routines. Without LDDMM, the deformation process fails to preserve surface smoothness and structural integrity. In contrast, the full model incorporating both LDDMM and scaling components (first row) produces smooth, realistic vascular surfaces that align well with underlying medical images. 

Quantitative evaluations further reinforce the visual observations. The rightmost two columns of Figure~\ref{fig:ablation_deform} illustrate the progression of misalignment losses and surface smoothness indicators across training epochs. In all configurations, the misalignment losses are relatively low, indicating overall good alignment with the underlying medical image data. Interestingly, both the ``no LDDMM'' and ``no Scaling'' cases exhibit lower misalignment losses compared to the full model. However, this reduction comes at a significant cost to geometric integrity, as evidenced by increased surface irregularities and unrealistic deformations in critical regions. These findings demonstrate that the optimization process tends to prioritize minimizing energy loss over preserving surface quality when structural constraints are absent. The inclusion of both LDDMM and scaling components is, therefore, essential for enforcing geometric fidelity and ensuring anatomically realistic surface reconstructions, balancing alignment accuracy with physical plausibility.

\section{Discussion}


This study presents a novel deep-learning framework that automates the construction of patient-specific vascular models for image-based CFD simulations, tackling long-standing challenges in traditional manual and semi-automated approaches. By integrating voxel-based segmentation through the LoGB-Net with a surface deformation module using GNN-LDDMM, the framework unifies traditionally separate workflows into a cohesive pipeline. This integration not only reduces the time and labor required for manual surface reconstruction but also minimizes operator variability, enhancing scalability and reproducibility. Importantly, the proposed approach demonstrates superior performance in both segmentation accuracy and mesh quality, as validated by quantitative metrics, visual assessments, and downstream CFD simulations.

One of the most notable advantages of the proposed framework is its ability to generate simulation-ready vascular geometries with minimal human intervention. Traditional manual workflows, such as those employed in tools like SimVascular, are labor-intensive, time-consuming, and highly dependent on operator expertise. These manual approaches often involve intricate tasks such as centerline extraction, cross-sectional segmentation, and surface interpolation, each prone to variability and subjective interpretation. In contrast, the proposed framework automates the whole process while ensuring anatomical accuracy. The surface deformation module further improves geometric fidelity by aligning reconstructed surfaces with underlying medical image gradients, eliminating artifacts commonly introduced by manual methods, such as oversmoothing or interpolation errors at junction regions.

While manual and automated geometries may appear visually similar, the results show that even minor geometric discrepancies can lead to substantial differences in simulated hemodynamic features such as pressure, wall shear stress, and velocity distributions. These parameters are critical biomarkers for cardiovascular disease diagnosis and treatment planning. Consequently, quantifying geometric uncertainty becomes critical to ensure the reliability of CFD-based analyses. The proposed framework incorporates uncertainty quantification (UQ) to assess segmentation variability and its propagation through surface deformation and CFD simulations. This capability enables the generation of an ensemble of plausible geometries and flow solutions, allowing users to better understand the range of potential outcomes.  
Our results demonstrate that uncertainty is significantly reduced in regions such as the main aorta through the image-guided surface deformation process, reflecting the robustness of the framework in achieving consistent and reliable results. However, persistent uncertainties in distal branches and areas with poor image quality highlight the influence of imaging limitations on segmentation performance.

\subsection{Limitation and future prospects of current framework}
Despite these strengths, the study has certain limitations that warrant further investigation. One limitation is the potential challenge faced by the LDDMM module when handling large deformations, such as those arising from poor initial segmentation results. In such cases, the deformation model may struggle to achieve accurate geometric reconstructions, particularly in highly distorted or irregular regions. This limitation underscores the need for more robust deformation methods or hybrid approaches to handle cases with significant discrepancies between the segmented geometry and the underlying image. 
Another limitation is the relatively small and homogeneous training dataset used to develop and evaluate the framework. While the results demonstrate great performance, expanding the dataset to include larger and more diverse samples is crucial for ensuring robustness across different patient populations, imaging conditions, and anatomical variations. This would enable the framework to better generalize to real-world clinical scenarios.
Furthermore, the robustness of LoGB-Net and GNN-LDDMM to severe image noise or poor-quality imaging data remains limited. For instance, distal branches and regions with low SNR often exhibit higher errors and poor robustness, which can propagate to CFD simulations. Enhancing the resilience of the framework to such challenges may involve additional anatomical priors, advanced noise-reduction techniques, or improved regularization strategies during the deformation process.

\subsection{Broader implications}

The automated and robust nature of the proposed framework has significant implications for advancing patient-specific cardiovascular modeling. By reducing the time, labor, and variability associated with manual workflows, this framework paves the way for integrating CFD-based analyses into routine clinical workflows. The incorporation of UQ further enhances the reliability of the results, enabling clinicians to make more informed decisions regarding diagnosis and treatment planning. Furthermore, the modular design of the framework allows for its extension to other anatomical regions and medical imaging modalities, broadening its applicability across various domains of computational medicine.

In conclusion, this study demonstrates the potential of combining deep learning with image-driven optimization to transform the field of patient-specific vascular modeling. By addressing critical limitations in existing methods and providing a comprehensive pipeline for automated model construction, the proposed framework represents a significant step toward making CFD-based analyses a practical and reliable tool in clinical practice. Future efforts will focus on addressing current limitations and exploring opportunities to enhance the scalability, robustness, and clinical impact of this approach.


\section{Materials and Methods}
\label{sec:meth}

\subsection{Dataset creation and preprocessing}
\label{sec:data}

This study leverages data from two open-source repositories: the ``Vascular Model Repository (VMR)'', curated by Wilson et al.\ \cite{Wilson2013}, and the ``Multicenter Aortic Vessel (AVT)'' dataset, released by Radl et al.\ \cite{Radl2022}. The VMR contains well-labeled, patient-specific segmentation samples of various cardiovascular structures, including the aorta, cerebral arteries, coronary arteries, aortofemoral arteries, and pulmonary arteries. Meanwhile, the AVT dataset comprises only aortic vessels with branches collected from three hospitals: KiTS, RIDER, and Dongyang. For this study, we focused on 32 aortic vessels with supra-aortic branches (e.g., right and left common carotid arteries and right and left subclavian arteries) from the VMR and an additional 16 samples from Dongyang’s hospital in the AVT dataset.

The VMR samples were originally labeled in the form of surface meshes created in SimVascular, which necessitated conversion to voxelized volume labels for our segmentation task. To achieve this, we developed a hole-filling algorithm that transforms surface meshes into binary voxel labels. For each voxel in the raw medical image, the algorithm identifies the nearest surface mesh point, computes its normal vector, and calculates the angle between this vector and the distance vector from the voxel to the surface point. Voxels with angles smaller than 90 degrees are classified as ``within the vessel'', while those exceeding this threshold are categorized as ``outside''. The resulting binary voxel labels maintain the same dimensions as the raw medical images, with one indicating ``within the vessel'' and zero indicating ``outside'', and are stored at the voxel centers.

To enhance segmentation robustness and ensure compatibility with the LoGB-Net model, the raw medical images underwent a series of preprocessing steps. (1) Resampling: The raw images were resampled using cubic interpolation, which increased the resolution by approximately $50\%$; (2) Clipping and normalization: Image intensity values were clipped between $0$ and $500$, then normalized to the range $[0, 1]$; (3) Background suppression: for training samples, pixel values outside the vessel regions (based on the image labels) were set to zero to focus the model's attention on vessel features. This step was omitted for testing samples, where image labels are assumed to be unavailable during inference.

To further bolster model generalizability, we applied data augmentation techniques, including random flipping, rotation, and cropping. The cropped dimensions were set to $64 \times 64 \times 64$, significantly smaller than the original image size of $512 \times 512 \times 128$. This approach not only increased the effective number of training samples but also conserved GPU memory during training. During inference, testing images were divided into overlapping cubic subregions of $64 \times 64 \times 64$, which were processed independently by the trained model to predict binary voxel labels. These predictions were subsequently stitched together to reconstruct the full segmentation of the original image. This pipeline ensures that the model can handle high-resolution inputs while maintaining computational efficiency.

\subsection{LoGB-Net: A multi-scale vascular segmentation model}
\label{sec:meth: LoGB}

\subsubsection{Hierarchical LoG kernels}

Our semantic segmentation model, Laplacian-of-Gaussian-based Bayesian network (LoGB-Net), features three primary components: a regular stream, a LoG stream, and an Atrous Spatial Pyramid Pooling (ASPP) module~\cite{Chen2018}, as illustrated in Figure~\ref{fig:deform}. 
The regular stream leverages a 3D CNN U-Net architecture, widely recognized for its efficacy in medical image segmentation tasks. This stream adopts a symmetrical encoder-decoder structure, where the encoder comprises five sequential blocks. Each block contains two $3 \times 3 \times 3$ convolutional layers, a Rectified Linear Unit (ReLU), and a $2 \times 2 \times 2$ max-pooling layer. To enhance feature fusion and mitigate vanishing gradient issues, skip connections are integrated between corresponding encoder and decoder layers, ensuring that high-resolution spatial features are preserved throughout the network.
The LoG stream enhances vessel detection across multiple scales by incorporating hierarchical 3D Laplacian-of-Gaussian (LoG) filters. These filters are mathematically defined as
\begin{equation}
        \nabla^2G(x,y,z)  = \frac{\partial^2G(x,y,z)}{\partial x^2} + \frac{\partial^2G(x,y,z)}{\partial y^2} + \frac{\partial^2G(x,y,z)}{\partial z^2},
\end{equation}
where the 3D Gaussian kernel $G(x, y, z)$ is expressed as
\begin{equation}
    G(x,y,z) = \frac{1}{2\pi\sigma^2}\exp\left[-\frac{x^2+y^2+z^2}{2\sigma^2}\right].
    \label{gaussian}
\end{equation}
Substituting $G(x, y, z)$ into the Laplacian operator results in the LoG filter
\begin{equation}
        \nabla^2G(x,y,z) =\frac{x^2+y^2+z^2-2\sigma^2}{\sigma^4}\exp\Big[-(x^2+y^2+z^2)/2\sigma^2\Big].
\end{equation}
Here, $(x, y, z)$ represents the 3D coordinates of the image voxel, and $\sigma$ controls the scale of the filter.
The rationale for combining the Laplacian and Gaussian operators lies in their complementary functionalities: the Laplacian operator excels at extracting second-order gradient information from the image, which is crucial for detecting edges and capturing structural details, while the Gaussian pre-filtering reduces the impact of noise by smoothing the image. This combination allows the model to retain the essential gradient features for segmentation while minimizing performance degradation caused by the noise sensitivity inherent to the Laplacian operation.

To capture the diverse dimensions of vascular structures in medical images, the LoG stream uses a hierarchical arrangement of filters. Previous studies have demonstrated that the kernel size directly correlates with the size of the target object~\cite{vermeer2004model}. Guided by this insight, we determined optimal kernel sizes of $3, 5, 7, 9, 11$ with corresponding $\sigma$ values of $0.5, 1.0, 1.5, 2.0, 2.5$. These five scales allow the LoGB-Net to effectively detect both large vessels, such as the main aorta, and smaller branches, like the carotid and subclavian arteries. During training, the kernel sizes and $\sigma$ values in LoGB-Net can be dynamically further optimized. This dynamic adaptation may enable the model to generalize across varying image resolutions and vessel structures.

\begin{figure}[t!]
    \centering
    \includegraphics[width=\textwidth]{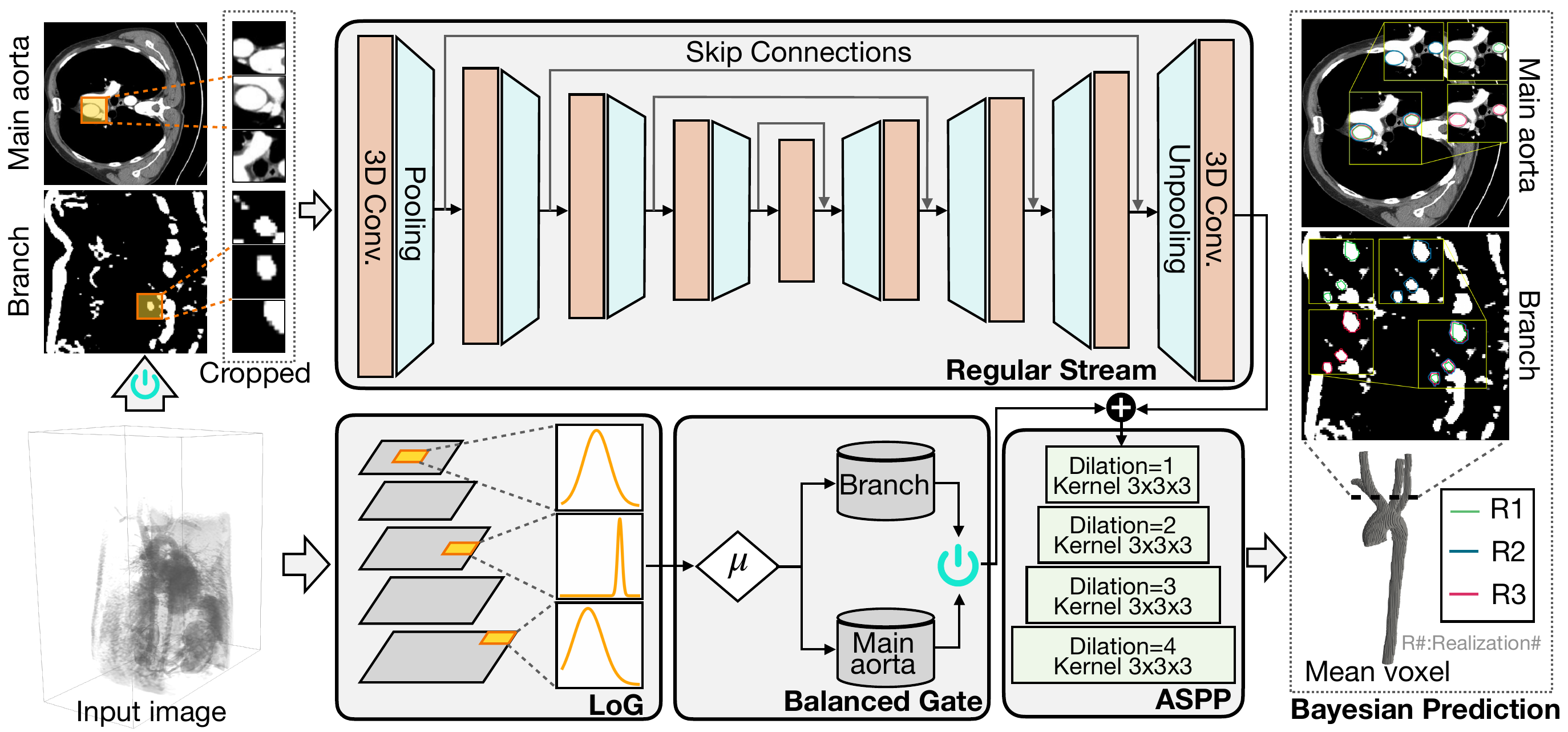}
    \caption{Overview of the Laplacian-of-Gaussian Bayesian Network (LoGB-Net) architecture for voxel-based segmentation of aortic vessels. Input medical images are cropped into main aorta and branch regions, processed through the regular stream and LoG stream for feature extraction. A balanced gate ensures equal focus on the main aorta and branch regions. Features are fused via an ASPP module, producing Bayesian predictions with mean segmentation and uncertainty quantification, capturing anatomically accurate structures.}
    \label{fig:segment}
\end{figure}

\subsubsection{The Bayesian settings}

Medical images often exhibit suboptimal signal quality, with blurred lumen edges and excessive noise. These artifacts make accurately delineating vessel boundaries particularly challenging. To address this issue, we formulate the training process of the LoG module in a Bayesian framework, which can estimate prediction uncertainty due to the presence of noise and data ambiguity, as inspired by similar approaches in Thiagarajan et al.\ \cite{Thiagarajan2022}.

In the Bayesian framework, the parameters $\theta$ of the LoG module are treated as random variables (RVs), and their posterior distribution given the data $D = \{(\mathrm{Image}, \mathrm{Label})_i\}_{i=1}^N$ is defined by Bayes' theorem
\begin{equation}
    p(\theta|D) = \frac{p(D|\theta)p(\theta)}{\int p(D|\theta)p(\theta)d\theta},
\end{equation}
where $p(\theta),p(D|\theta),p(\theta|D)$ represents the prior, likelihood and the posterior, respectively. In practice, directly computing the posterior is intractable due to the high dimensionality of the trainable parameter space. Therefore, we approximate the posterior using mean-field variational inference by assuming that the trainable parameters $\theta$ are Gaussian RVs. Namely, the prior distributions $p(\theta)$ are modeled as standard Gaussian, and the posterior distributions are also Gaussian, parameterized by mean and variances. The variational approximation is optimized by maximizing the Evidence Lower Bound (ELBO)
\begin{equation}
    \mathrm{ELBO} = \mathrm{E}_{q(\theta|D)}[\log p(D|\theta)] - \mathrm{KL}(q(\theta|D)||p(\theta)),
\end{equation}
where the first term is the expected log-likelihood, measuring how well the model explains the data, and the second term evaluates the Kullback-Leibler divergence, penalizing the deviations of the variational distribution $q(\theta|D)$ from the prior $p(\theta)$. We use the reparameterization trick to sample weights $\theta$, which ensures that the sampling process is differentiable with respect to the ELBO objective, enabling gradient-based optimization.  

In addition to the Bayesian ELBO, we incorporate the Dice coefficient loss to optimize segmentation accuracy. The Dice coefficient measures the overlap between the predicted segmentation and the ground truth and is defined as
\begin{equation}
    \mathrm{Dice} = \frac{2\sum_{i=1}^{N_A}g_io_i}{\sum_{i=1}^{N_A}g_i^2+\sum_{i=1}^{N_A}o_i^2},
\label{eq:dice}
\end{equation}
where $g_i$ and $o_i$ represent the ground truth and predicted voxel values, respectively. $N_A$ denotes the total number of voxels. The Dice loss is expressed as: $\mathcal{L}_\mathrm{Dice} = 1 - \mathrm{Dice}$. The training process involves optimizing a combined loss function that maximizes the ELBO and minimizes the Dice loss
\begin{equation}
    \hat\theta =  \arg\min_{\theta} \left( \mathcal{L}_\mathrm{Dice} - \mathrm{ELBO} \right).
    \label{overall_loss}
\end{equation}
To incorporate domain-specific knowledge, we initialize the posterior mean of the Bayesian convolutional layers with precomputed Laplacian of Gaussian (LoG) kernels. This initialization biases the model toward detecting vascular structures, facilitating faster convergence and improved performance in early training stages.

\subsubsection{The balanced gate} 
During training, medical images are randomly cropped into ($64 \times 64\times 64$) to standardize input dimensions and optimize computational efficiency. However, this approach leads to a significant imbalance in the distribution of training samples, as cubes containing the main aorta vessel are far more prevalent than those containing smaller vascular branches. This imbalance occurs because the main aorta occupies a larger volume in the imaging domain, while smaller branches, such as the supra-aortic vessels, are comparatively limited in size and representation. Without addressing this imbalance, the model tends to prioritize the detection of the main aorta, resulting in suboptimal performance on smaller branches, as observed in preliminary experiments.

To mitigate this issue, a balanced gate was incorporated adjacent to the LoG stream to ensure equitable representation of both large and small vessel regions during training. This mechanism leverages the output of the LoG stream, which is specifically designed to highlight vessel edges. For each input cube, a voxel percentage metric $V_p$ is computed, defined as the proportion of voxels in the cube that exceed a threshold value of 0.5
\begin{equation}
    V_p = \frac{\sum_{i=1}^{N_c} \mathbf{1}(o_i > 0.5)}{N_c},
\end{equation}
where $v_i$ represents the predicted value of the $i$-th voxel in the cube, $N_c$ is the total number of voxels in the cube, and $\mathbf{1}(\cdot)$ is an indicator function that returns one if the condition is satisfied and zero otherwise. Since the LoG kernels are optimized for edge detection, a higher voxel percentage indicates a higher likelihood of the cube containing larger vessels, such as the main aorta. Conversely, a lower voxel percentage is indicative of smaller branches or regions with fewer vessel features.

Using a threshold value of $\mu = 15\%$, the balanced gate categorizes cubes into two groups: cubes with voxel percentages exceeding the threshold are labeled as ``large vessels'', while those below the threshold are categorized as ``small branches''. During training, the data preprocessing algorithm feeds input cubes into the balanced gate, which dynamically selects an equal number of samples from each group. Specifically, the algorithm selects five cubes from each group to construct a balanced batch of ten samples. Once the batch quota is reached, it is finalized, and the training step is initialized. The balanced gate mechanism addresses the inherent data imbalance, ensuring that the model is exposed to sufficient training examples of both large and small vascular structures. This approach is particularly critical for achieving high segmentation accuracy across multi-branch vascular geometries.


\subsubsection{Training, testing, and evaluation metrics}
\label{sec:train and test}

The training and testing of our model, along with other baseline models, were conducted using a combination of two open-source datasets: VMR and AVT. The total dataset was divided into $66\%$ for training and $33\%$ for testing. Specifically, 24 samples were taken from VMR, and eight samples were obtained from AVT for training purposes. For testing, 16 samples were reserved and evenly distributed between the two data sources.  
All training experiments were conducted on an NVIDIA GeForce RTX 3080 with 12GB of memory. Each model was trained for 5000 epochs, requiring approximately 18 hours. The training process utilized a learning rate of $1\times10^{-5}$ and a batch size of 8, optimized using the Adam optimizer. 

To assess model performance, we employed three widely recognized metrics in the image segmentation community: Dice coefficient, Average Surface Distance (ASD), and Hausdorff distance. The Dice coefficient, mathematically expressed in Equation~\ref{eq:dice}, quantifies the overlap between the predicted segmentation and the corresponding ground truth.
The ASD measures the mean distance between the surface point cloud of predicted segmentation ($S_1$) and the ground truth ($S_2$), defined as follows
\begin{equation}
    ASD(S_1, S_2) = \frac{1}{|S_1| + |S_2|} \left( \sum_{x \in S_1} \min_{y \in S_2} d(x, y) + \sum_{y \in S_2} \min_{x \in S_1} d(y, x) \right),
    \label{eq:asd}
\end{equation}
where $|S_1|$ and $|S_2|$ represent the number of surface points in the predicted surface and the ground truth segmentations, respectively, and $d(x,y)$ denotes the Euclidean distance between points $x$ and $y$.
The Hausdorff distance evaluates the maximum surface discrepancy by identifying the largest minimal distance between points on the predicted and ground truth surfaces. It is defined as
\begin{equation}
H(X, Y) = \max \left\{ \sup_{x \in S_1} \inf_{y \in S_1} d(x, y), \sup_{y \in S_2} \inf_{x \in S_1} d(x, y) \right\},
\end{equation}
where $\sup$ and $\inf$ denote the supremum and infimum, respectively. 
For an optimal segmentation model, the Dice coefficient should approach 1, indicating near-perfect overlap between the predicted and ground truth segmentations. Conversely, smaller values for ASD and Hausdorff distance indicate improved geometric fidelity and minimized surface discrepancies. These metrics collectively provide a robust evaluation framework for comparing the performance of different models on vascular segmentation tasks.

\subsection{The GNN-LDDMM framework for surface reconstruction and refinement}

\subsubsection{Preprocessing}
The preprocessing module begins by reconstructing a triangular surface mesh from the predicted binary voxel image from the LoGB-Net. This is accomplished by extracting the isosurface using the Marching Cubes algorithm~\cite{lorensen1998marching}, a widely used technique for generating 3D surface representations from volumetric data. However, the initial mesh produced by this method often exhibits irregular vertex distributions and a ``LEGO-like'' blocky appearance due to the voxelized nature of the input. To address these issues, the mesh is resampled using the Approximated Centroidal Voronoi Diagrams (ACVD) algorithm~\cite{valette2004approximated}, which redistributes vertices to achieve a more uniform and structured layout. This resampling step not only enhances the visual quality of the mesh but also improves its geometric consistency, making it more amenable to subsequent processing stages. To further eliminate the blocky surface, the resampled mesh undergoes a smoothing process implemented with the PyTorch3D package~\cite{ravi2020accelerating}. This process involves minimizing three regularization terms: normal loss, which aligns neighboring surface normals to reduce abrupt changes in orientation; edge loss, which enforces uniformity in edge lengths to prevent distortions; and Laplacian loss, which smooths the surface by minimizing local curvature variations. Together, these operations produce a refined and well-contoured mesh that preserves anatomical fidelity and is optimized for subsequent deformation processes.

\subsubsection{GNN-LDDMM surface deformation module}
\begin{figure}[t!]
    \centering
    \includegraphics[width=\textwidth]{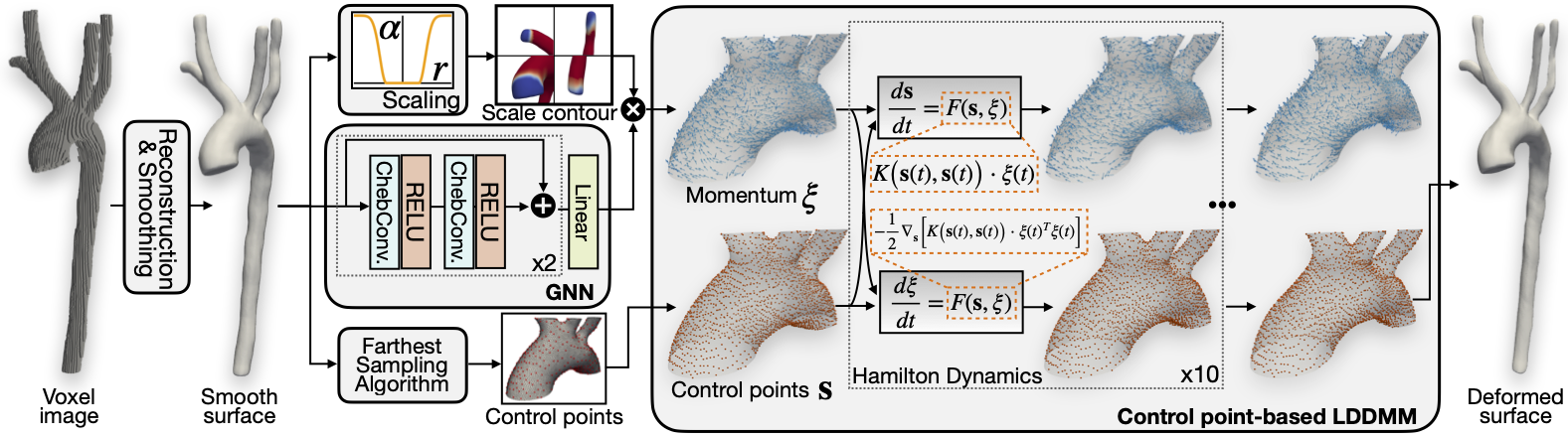}
    \caption{The GNN-LDDMM module refines the surface mesh using a Graph Neural Network (GNN) to predict the initial momentum field, processed through Chebyshev convolution layers. The momentum field drives the Large Deformation Diffeomorphic Metric Mapping (LDDMM) framework, which iteratively deforms the surface via Hamiltonian dynamics, ensuring smooth and anatomically accurate transformations.}
    \label{fig:deform}
\end{figure}

The GNN-LDDMM surface deformation module is designed to refine the alignment of the reconstructed surface mesh with the underlying image background in an unsupervised manner. This module integrates a graph neural network (GNN) to predict a vector field on the surface point cloud and a large deformation diffeomorphic metric mapping (LDDMM) algorithm to ensure smooth, anatomically consistent deformations (see Figure~\ref{fig:deform}). Inspired by~\cite{beg2005computing}, the LDDMM component is specifically tailored to avoid surface distortions and inaccuracies, which are common in GNN-based deformation approaches.

The GNN component consists of two blocks, each incorporating two Chebyshev convolutional (ChevConv) layers with residual connections. These layers process the point coordinates of the surface mesh and compute a vector field $v$ over the surface point cloud. This vector field serves as the initial momentum for the subsequent LDDMM-based deformation.

The LDDMM operates on the initial surface \( \mathbf{S} \), which is uniformly subsampled to generate a set of control points \( \{\mathbf{s}_i\}_{i=1}^{N_s} \) as detailed in ~\cite{qi2017pointnet++}. Each control point is initialized with a momentum vector \( \{\bm{\xi}\}_{i=1}^{N_s} \), and together these define a velocity field through kernel interpolation
\begin{equation}
v(x) = \sum_{i=1}^{N_s} K(\mathbf{x}, \mathbf{s}_i) \cdot \bm{\xi}_i,
\label{eq:v}
\end{equation}
where $K$ is the standard Gaussian kernel
\begin{equation}
    K(\mathbf{x},\mathbf{y}) = \exp\left(-\frac{||\mathbf{x}-\mathbf{y}||^2}{\sigma^2}\right).
\end{equation}
The control points and their associated momenta evolve iteratively under Hamiltonian dynamics
\begin{equation}
\begin{cases}
\begin{aligned}
    \frac{d\mathbf{s}}{dt} &= K\big(\mathbf{s}(t), \mathbf{s}(t)\big) \cdot \bm{\xi}(t) \\
    \frac{d\bm{\xi}}{dt} &= -\frac{1}{2} \nabla_\mathbf{s} \Big[ K\big(\mathbf{s}(t), \mathbf{s}(t)\big) \cdot \bm{\xi}(t)^T \bm{\xi}(t) \Big]. 
\end{aligned}
\end{cases}
\end{equation}
These equations are numerically integrated using a second-order Runge-Kutta method to compute the time-dependent velocity field $\mathbf{v}(\mathbf{x},t)$. This velocity field governs the deformation of the surface over a discretized time span \( T = 1 \), divided into 15 steps. The resulting movement of the surface is described by the following ordinary differential equation (ODE)
\begin{equation}
\frac{d\mathbf{S}}{dt} = \mathbf{v}(\mathbf{S},t).
\end{equation}
The deformed surface $\hat{\mathbf{S}}$ is obtained by integrating the ODE over the entire time span
\begin{equation}
\hat{\mathbf{S}} = \Phi\left(\mathbf{S};\mathbf{s},\bm{\xi} \right) = \int_{t=0}^1 \mathbf{v}_{\mathbf{s},\bm{\xi}}( \mathbf{S}, t)dt, 
\end{equation}
where $\Phi(\mathbf{S};\mathbf{s},\bm{\xi})$ represents the diffeomorphic transformation, fully parameterized by the initial control points $\mathbf{s}$ and momentum $\bm{\xi}$.

The integration of the GNN and LDDMM components ensures that the deformation process is both data-driven and anatomically consistent. The GNN generates an initial momentum field that adapts to the specific geometry of the surface, while the LDDMM guarantees that the resulting transformations maintain topological consistency. This combination is particularly effective for handling complex vascular structures, where traditional deformation methods often introduce distortions or fail to align surfaces accurately.

\subsubsection{The scaling gate}
One of the major challenges during the deformation process is the severe distortion observed at the inlets and outlets of the aorta. These regions are not the primary focus of the alignment process, which primarily targets the vessel wall, leading to control points near the inlets and outlets being optimized to arbitrary locations. Such behavior often results in undesirable deformations at these boundary regions. To mitigate this issue, we implemented a scaling gate to restrict the movement of surface points, specifically at the inlets and outlets.

Before initiating the deformation, a scalar field $\{\alpha(\mathbf{x}_i)\}_{i=1}^{N_s}$ is computed over the control point cloud. This scalar field determines the degree of mobility for each surface point based on its location relative to the inlets and outlets. Specifically, for points located within the inlets or outlets, the geodesic distance from the center point to the boundary (geodesic radius) is calculated. Then the scalar field $\alpha{\mathbf{x}}$ is defined as
\begin{equation}
\alpha(x) = 
\begin{cases}
0, \;\;\;\;\;\;\;\;\;\;\;\;\;\;\;\;\;\;\;\;\;\;\;\;\;\;\;\;\;\;\;\;\;\;\;\;\;\;\;|\mathbf{x}-\mathbf{x}^i_c|<r^i\;\;\forall i\in[0,N_{I/O}],\\
1-exp\left(-\frac{(|\mathbf{x}-\mathbf{x}^i_c|-r^i)^2}{2\sigma^2}\right), \;\;\;\;\;\;r^i\leq |\mathbf{x}-\mathbf{x}^i_c|<r^i+3\sigma\;\;\forall i\in[0,N_{I/O}],\\
1, \;\;\;\;\;\;\;\;\;\;\;\;\;\;\;\;\;\;\;\;\;\;\;\;\;\;\;\;\;\;\;\;\;\;\;\;\;\;\;|\mathbf{x}-\mathbf{x}^i_c|\geq r^i+3\sigma \;\;\forall i\in[0,N_{I/O}],
\end{cases}
\end{equation}
where $\{\mathbf{x}_c^i\}_{i=1}^{N_{I/O}}$ are the center points of the inlets and outlets, $\{r^i\}_{i=1}^{N_{I/O}}$ are the geodesic radii of the inlets and outlets, ${N_{I/O}}$ is the total number of inlets and outlets in the vascular tree, and $\sigma$ controls the width of the buffer zone, ensuring a smooth transition between fully immobilized and fully mobile regions. Namely, for points on the inlets or outlets, $\alpha = 0$, effectively immobilizing these points; for points along the vessel wall, $\alpha = 1$, allowing full mobility; for points in the transitional region, $\alpha$ smoothly transitions from 1 to 0 using a truncated Gaussian function to ensure a gradual gradation in mobility constraints. 

\subsubsection{Unsupervised training based on input medical imaging}

The training process for the surface deformation module is designed to operate in an unsupervised manner, relying solely on the input medical imaging data to drive the alignment and ensure high-quality mesh deformation. The objective function consists of two main components: an energy term that quantifies the alignment between the deformed surface and the input medical image and surface regularization terms that ensure the geometric quality of the resulting mesh. The total loss function is formulated as
\begin{equation}
\mathcal{L}_\mathrm{total} =  - w_1 \log\left(\sum^{N_A}_i\mathcal{G}(s_i)\right) + w_2 \mathcal{L}_\mathrm{normal} + w_3 \mathcal{L}_\mathrm{edge} + w_4 \mathcal{L}_\mathrm{laplacian}. 
\end{equation}
where $\mathcal{G}(s_i)$ represents the image gradient magnitude at the $i$-th surface point $s_i$, and $N_A$ is the total number of points on the surface mesh. The terms $\mathcal{L}_{\text{normal}}$, $\mathcal{L}_{\text{edge}}$, and $\mathcal{L}_{\text{laplacian}}$ are the surface regularization terms, and \( w_1, w_2, w_3, w_4 \) are the associated weights.

The first term, referred to as the misalignment energy, is defined as the negative logarithm of the sum of image gradient magnitudes over the surface point cloud. Minimizing this term encourages the surface points to move towards regions in the medical image with the largest gradients, corresponding to anatomical boundaries or regions of high contrast. This ensures that the deformed surface aligns accurately with the image data, capturing the underlying anatomical structure.

The second term, $\mathcal{L}_{\text{normal}}$, referred to as the normal loss, penalizes deviations in the orientation of adjacent surface normals. By encouraging parallelism between neighboring normals, this term prevents abrupt changes in surface orientation and mitigates the risk of folding or kinks in the mesh
\begin{equation}
\mathcal{L}_{\text{normal}} = \frac{1}{N_{IE}} \sum_{i=1}^{N_{IE}} \left(1 - \mathbf{n}_1 \cdot \mathbf{n}_2 \right),
\end{equation}
where $N_{IE}$ is the total number of internal edges on a triangulated mesh, and  $\mathbf{n}_1$ and $\mathbf{n}_2$ are the surface normals of the two neighboring triangular elements $e_i$ and $e_j$.

The third term, $\mathcal{L}_{\text{edge}}$, is the edge loss, which ensures uniformity in edge lengths across the mesh. This term is defined as
\begin{equation}
\mathcal{L}_{\text{edge}} = \frac{1}{N_E} \sum_{e \in \mathcal{E}} \left(\| \mathbf{p}_e \| - \bar{l}\right)^2,
\end{equation}
where $\mathcal{E}$ is the set of all edges in the mesh, $\mathbf{p}_e$ represents the edge vector, $\bar{l}$ is the average edge length, and $N_E$ is the total number of edges. Minimizing this term reduces regional distortions and improves the uniformity of the mesh geometry.

The last term, the Laplacian loss $\mathcal{L}_{\text{laplacian}}$~\cite{nealen2006laplacian}, enhances the smoothness of the surface by penalizing deviations from local surface curvature. It is expressed as
\begin{equation}
\mathcal{L}_{\text{laplacian}} = \frac{1}{N_A} \sum_{i=1}^{N_A} \| \mathbf{s}_i - \frac{1}{|\mathcal{N}(i)|} \sum_{j \in \mathcal{N}(i)} \mathbf{s}_j \|^2,
\end{equation}
where $\mathbf{s}_i$ is the position of the $i$-th surface point, and $\mathcal{N}(i)$ is its set of neighbors. This term smooths the surface by ensuring that each point’s position aligns closely with the average position of its neighbors, thereby reducing sharp irregularities.

For all samples, the weights of the loss terms are set to $w_1 = 1.0$, $w_2 = 0.2$, $w_3 = 0.01$, and $w_4 = 0.1$. These values are chosen to balance alignment with the image background and the geometric quality of the surface mesh. By minimizing $\mathcal{L}_{\text{total}}$, the optimization process achieves accurate surface alignment while maintaining a high-quality mesh that is ready for downstream CFD/FSI simulations.



\section*{Acknowledgment}
The authors would like to acknowledge funding support from the National Science Foundation under award numbers OAC-2047127 and OAC-2104158 and the National Institutes of Health under award number 1R01HL177814.

\section*{Data avilibility}
All data needed to evaluate the conclusions in the paper are present in the paper and/or the Supplementary Materials.

\section*{Compliance with Ethical Standards}
Conflict of Interest: The authors declare that they have no conflict of interest.





\begin{thebibliography}{10}
    \expandafter\ifx\csname url\endcsname\relax
      \def\url#1{\texttt{#1}}\fi
    \expandafter\ifx\csname urlprefix\endcsname\relax\def\urlprefix{URL }\fi
    \expandafter\ifx\csname href\endcsname\relax
      \def\href#1#2{#2} \def\path#1{#1}\fi
    
    \bibitem{csahin2022risk}
    B.~{\c{S}}ahin, G.~{\.I}lg{\"u}n, Risk factors of deaths related to
      cardiovascular diseases in world health organization (who) member countries,
      Health \& Social Care in the Community 30~(1) (2022) 73--80.
    
    \bibitem{steinman2002image}
    D.~A. Steinman, Image-based computational fluid dynamics modeling in realistic
      arterial geometries, Annals of biomedical engineering 30 (2002) 483--497.
    
    \bibitem{taylor2009patient}
    C.~A. Taylor, C.~Figueroa, Patient-specific modeling of cardiovascular
      mechanics, Annual review of biomedical engineering 11~(1) (2009) 109--134.
    
    \bibitem{gray2018patient}
    R.~A. Gray, P.~Pathmanathan, Patient-specific cardiovascular computational
      modeling: diversity of personalization and challenges, Journal of
      cardiovascular translational research 11 (2018) 80--88.
    
    \bibitem{updegrove2017simvascular}
    A.~Updegrove, N.~M. Wilson, J.~Merkow, H.~Lan, A.~L. Marsden, S.~C. Shadden,
      Simvascular: an open source pipeline for cardiovascular simulation, Annals of
      biomedical engineering 45 (2017) 525--541.
    
    \bibitem{du2022deep}
    P.~Du, X.~Zhu, J.-X. Wang, Deep learning-based surrogate model for
      three-dimensional patient-specific computational fluid dynamics, Physics of
      Fluids 34~(8) (2022) 081906.
    
    \bibitem{arzani2022machine}
    A.~Arzani, J.-X. Wang, M.~S. Sacks, S.~C. Shadden, Machine learning for
      cardiovascular biomechanics modeling: challenges and beyond, Annals of
      Biomedical Engineering 50~(6) (2022) 615--627.
    
    \bibitem{du2022reducing}
    P.~Du, J.-X. Wang, Reducing geometric uncertainty in computational hemodynamics
      by deep learning-assisted parallel-chain mcmc, Journal of Biomechanical
      Engineering 144~(12) (2022) 121009.
    
    \bibitem{ajam2017review}
    A.~Ajam, A.~A. Aziz, V.~S. Asirvadam, A.~S. Muda, I.~Faye, S.~J.~S. Gardezi, A
      review on segmentation and modeling of cerebral vasculature for surgical
      planning, IEEE Access 5 (2017) 15222--15240.
    
    \bibitem{rueckert1997automatic}
    D.~Rueckert, P.~Burger, S.~Forbat, R.~Mohiaddin, G.-Z. Yang, Automatic tracking
      of the aorta in cardiovascular mr images using deformable models, IEEE
      Transactions on medical imaging 16~(5) (1997) 581--590.
    
    \bibitem{das2006aortic}
    B.~Das, Y.~Mallya, S.~Srikanth, R.~Malladi, Aortic thrombus segmentation using
      narrow band active contour model, in: 2006 International Conference of the
      IEEE Engineering in Medicine and Biology Society, IEEE, 2006, pp. 408--411.
    
    \bibitem{krissian2014semi}
    K.~Krissian, J.~M. Carreira, J.~Esclarin, M.~Maynar, Semi-automatic
      segmentation and detection of aorta dissection wall in mdct angiography,
      Medical image analysis 18~(1) (2014) 83--102.
    
    \bibitem{wang2017segmentation}
    Y.~Wang, F.~Seguro, E.~Kao, Y.~Zhang, F.~Faraji, C.~Zhu, H.~Haraldsson,
      M.~Hope, D.~Saloner, J.~Liu, Segmentation of lumen and outer wall of
      abdominal aortic aneurysms from 3d black-blood mri with a registration based
      geodesic active contour model, Medical image analysis 40 (2017) 1--10.
    
    \bibitem{ling2019fast}
    H.~Ling, J.~Gao, A.~Kar, W.~Chen, S.~Fidler, Fast interactive object annotation
      with curve-gcn, in: Proceedings of the IEEE/CVF conference on computer vision
      and pattern recognition, 2019, pp. 5257--5266.
    
    \bibitem{leventon2002statistical}
    M.~E. Leventon, W.~E.~L. Grimson, O.~Faugeras, Statistical shape influence in
      geodesic active contours, in: 5th IEEE EMBS International Summer School on
      Biomedical Imaging, 2002., IEEE, 2002, pp. 8--pp.
    
    \bibitem{he2008comparative}
    L.~He, Z.~Peng, B.~Everding, X.~Wang, C.~Y. Han, K.~L. Weiss, W.~G. Wee, A
      comparative study of deformable contour methods on medical image
      segmentation, Image and vision computing 26~(2) (2008) 141--163.
    
    \bibitem{jorstad2015neuromorph}
    A.~Jorstad, B.~Nigro, C.~Cali, M.~Wawrzyniak, P.~Fua, G.~Knott, Neuromorph: a
      toolset for the morphometric analysis and visualization of 3d models derived
      from electron microscopy image stacks, Neuroinformatics 13 (2015) 83--92.
    
    \bibitem{mcinerney1995dynamic}
    T.~McInerney, D.~Terzopoulos, A dynamic finite element surface model for
      segmentation and tracking in multidimensional medical images with application
      to cardiac 4d image analysis, Computerized medical imaging and graphics
      19~(1) (1995) 69--83.
    
    \bibitem{terzopoulos1988constraints}
    D.~Terzopoulos, A.~Witkin, M.~Kass, Constraints on deformable models:
      Recovering 3d shape and nonrigid motion, Artificial intelligence 36~(1)
      (1988) 91--123.
    
    \bibitem{terzopoulos1988symmetry}
    D.~Terzopoulos, A.~Witkin, M.~Kass, Symmetry-seeking models and 3d object
      reconstruction, International Journal of Computer Vision 1~(3) (1988)
      211--221.
    
    \bibitem{lareyre2019fully}
    F.~Lareyre, C.~Adam, M.~Carrier, C.~Dommerc, C.~Mialhe, J.~Raffort, A fully
      automated pipeline for mining abdominal aortic aneurysm using image
      segmentation, Scientific reports 9~(1) (2019) 13750.
    
    \bibitem{bidhult2019new}
    S.~Bidhult, E.~Hedstr{\"o}m, M.~Carlsson, J.~T{\"o}ger, K.~Steding-Ehrenborg,
      H.~Arheden, A.~H. Aletras, E.~Heiberg, A new vessel segmentation algorithm
      for robust blood flow quantification from two-dimensional phase-contrast
      magnetic resonance images, Clinical physiology and functional imaging 39~(5)
      (2019) 327--338.
    
    \bibitem{antiga2008image}
    L.~Antiga, M.~Piccinelli, L.~Botti, B.~Ene-Iordache, A.~Remuzzi, D.~A.
      Steinman, An image-based modeling framework for patient-specific
      computational hemodynamics, Medical \& biological engineering \& computing 46
      (2008) 1097--1112.
    
    \bibitem{osher2004level}
    S.~Osher, R.~Fedkiw, K.~Piechor, Level set methods and dynamic implicit
      surfaces, Appl. Mech. Rev. 57~(3) (2004) B15--B15.
    
    \bibitem{volonghi2016automatic}
    P.~Volonghi, D.~Tresoldi, M.~Cadioli, A.~M. Usuelli, R.~Ponzini, U.~Morbiducci,
      A.~Esposito, G.~Rizzo, Automatic extraction of three-dimensional thoracic
      aorta geometric model from phase contrast mri for morphometric and
      hemodynamic characterization, Magnetic Resonance in Medicine 75~(2) (2016)
      873--882.
    
    \bibitem{kurugol2015automated}
    S.~Kurugol, C.~E. Come, A.~A. Diaz, J.~C. Ross, G.~L. Kinney, J.~L.
      Black-Shinn, J.~E. Hokanson, M.~J. Budoff, G.~R. Washko, R.~San Jose~Estepar,
      Automated quantitative 3d analysis of aorta size, morphology, and mural
      calcification distributions, Medical physics 42~(9) (2015) 5467--5478.
    
    \bibitem{zhuge2006abdominal}
    F.~Zhuge, G.~D. Rubin, S.~Sun, S.~Napel, An abdominal aortic aneurysm
      segmentation method: Level set with region and statistical information,
      Medical physics 33~(5) (2006) 1440--1453.
    
    \bibitem{chen2020deep}
    C.~Chen, C.~Qin, H.~Qiu, G.~Tarroni, J.~Duan, W.~Bai, D.~Rueckert, Deep
      learning for cardiac image segmentation: a review, Frontiers in
      Cardiovascular Medicine 7 (2020) 25.
    
    \bibitem{jia2021learning}
    D.~Jia, X.~Zhuang, Learning-based algorithms for vessel tracking: A review,
      Computerized Medical Imaging and Graphics 89 (2021) 101840.
    
    \bibitem{wolterink2016dilated}
    J.~M. Wolterink, T.~Leiner, M.~A. Viergever, I.~I{\v{s}}gum, Dilated
      convolutional neural networks for cardiovascular mr segmentation in
      congenital heart disease, in: International Workshop on Reconstruction and
      Analysis of Moving Body Organs, Springer, 2016, pp. 95--102.
    
    \bibitem{bai2018recurrent}
    W.~Bai, H.~Suzuki, C.~Qin, G.~Tarroni, O.~Oktay, P.~M. Matthews, D.~Rueckert,
      Recurrent neural networks for aortic image sequence segmentation with sparse
      annotations, in: Medical Image Computing and Computer Assisted
      Intervention--MICCAI 2018: 21st International Conference, Granada, Spain,
      September 16-20, 2018, Proceedings, Part IV 11, Springer, 2018, pp. 586--594.
    
    \bibitem{xia2019automatic}
    Q.~Xia, Y.~Yao, Z.~Hu, A.~Hao, Automatic 3d atrial segmentation from ge-mris
      using volumetric fully convolutional networks, in: Statistical Atlases and
      Computational Models of the Heart. Atrial Segmentation and LV Quantification
      Challenges: 9th International Workshop, STACOM 2018, Held in Conjunction with
      MICCAI 2018, Granada, Spain, September 16, 2018, Revised Selected Papers 9,
      Springer, 2019, pp. 211--220.
    
    \bibitem{vigneault2018omega}
    D.~M. Vigneault, W.~Xie, C.~Y. Ho, D.~A. Bluemke, J.~A. Noble, $\omega$-net
      (omega-net): fully automatic, multi-view cardiac mr detection, orientation,
      and segmentation with deep neural networks, Medical image analysis 48 (2018)
      95--106.
    
    \bibitem{wolterink2018automatic}
    J.~M. Wolterink, T.~Leiner, M.~A. Viergever, I.~I{\v{s}}gum, Automatic
      segmentation and disease classification using cardiac cine mr images, in:
      Statistical Atlases and Computational Models of the Heart. ACDC and MMWHS
      Challenges: 8th International Workshop, STACOM 2017, Held in Conjunction with
      MICCAI 2017, Quebec City, Canada, September 10-14, 2017, Revised Selected
      Papers 8, Springer, 2018, pp. 101--110.
    
    \bibitem{baumgartner2018exploration}
    C.~F. Baumgartner, L.~M. Koch, M.~Pollefeys, E.~Konukoglu, An exploration of 2d
      and 3d deep learning techniques for cardiac mr image segmentation, in:
      Statistical Atlases and Computational Models of the Heart. ACDC and MMWHS
      Challenges: 8th International Workshop, STACOM 2017, Held in Conjunction with
      MICCAI 2017, Quebec City, Canada, September 10-14, 2017, Revised Selected
      Papers 8, Springer, 2018, pp. 111--119.
    
    \bibitem{newman2006survey}
    T.~S. Newman, H.~Yi, A survey of the marching cubes algorithm, Computers \&
      Graphics 30~(5) (2006) 854--879.
    
    \bibitem{zhao2022segmentation}
    J.~Zhao, J.~Zhao, S.~Pang, Q.~Feng, Segmentation of the true lumen of aorta
      dissection via morphology-constrained stepwise deep mesh regression, IEEE
      Transactions on Medical Imaging 41~(7) (2022) 1826--1836.
    
    \bibitem{wickramasinghe2021deep}
    U.~Wickramasinghe, P.~Fua, G.~Knott, Deep active surface models, in:
      Proceedings of the IEEE/CVF Conference on Computer Vision and Pattern
      Recognition, 2021, pp. 11652--11661.
    
    \bibitem{kong2021deep}
    F.~Kong, N.~Wilson, S.~Shadden, A deep-learning approach for direct whole-heart
      mesh reconstruction, Medical image analysis 74 (2021) 102222.
    
    \bibitem{bongratz2022vox2cortex}
    F.~Bongratz, A.-M. Rickmann, S.~P{\"o}lsterl, C.~Wachinger, Vox2cortex: fast
      explicit reconstruction of cortical surfaces from 3d mri scans with geometric
      deep neural networks, in: Proceedings of the IEEE/CVF Conference on Computer
      Vision and Pattern Recognition, 2022, pp. 20773--20783.
    
    \bibitem{wickramasinghe2020voxel2mesh}
    U.~Wickramasinghe, E.~Remelli, G.~Knott, P.~Fua, Voxel2mesh: 3d mesh model
      generation from volumetric data, in: Medical Image Computing and Computer
      Assisted Intervention--MICCAI 2020: 23rd International Conference, Lima,
      Peru, October 4--8, 2020, Proceedings, Part IV 23, Springer, 2020, pp.
      299--308.
    
    \bibitem{deng2022survey}
    B.~Deng, Y.~Yao, R.~M. Dyke, J.~Zhang, A survey of non-rigid 3d registration,
      Computer Graphics Forum 41~(2) (2022) 559--589.
    
    \bibitem{eisenberger2021neuromorph}
    M.~Eisenberger, D.~Novotny, G.~Kerchenbaum, P.~Labatut, N.~Neverova,
      D.~Cremers, A.~Vedaldi, Neuromorph: Unsupervised shape interpolation and
      correspondence in one go, in: Proceedings of the IEEE/CVF Conference on
      Computer Vision and Pattern Recognition, 2021, pp. 7473--7483.
    
    \bibitem{amor2022resnet}
    B.~B. Amor, S.~Arguill{\`e}re, L.~Shao, Resnet-lddmm: advancing the lddmm
      framework using deep residual networks, IEEE Transactions on Pattern Analysis
      and Machine Intelligence 45~(3) (2022) 3707--3720.
    
    \bibitem{eisenberger2020smooth}
    M.~Eisenberger, Z.~Lahner, D.~Cremers, Smooth shells: Multi-scale shape
      registration with functional maps, in: Proceedings of the IEEE/CVF Conference
      on Computer Vision and Pattern Recognition, 2020, pp. 12265--12274.
    
    \bibitem{beg2005computing}
    M.~F. Beg, M.~I. Miller, A.~Trouv{\'e}, L.~Younes, Computing large deformation
      metric mappings via geodesic flows of diffeomorphisms, International journal
      of computer vision 61 (2005) 139--157.
    
    \bibitem{updegrove2016boolean}
    A.~Updegrove, N.~M. Wilson, S.~C. Shadden, Boolean and smoothing of discrete
      polygonal surfaces, Advances in Engineering Software 95 (2016) 16--27.
    
    \bibitem{Lin_2017_CVPR}
    T.-Y. Lin, P.~Dollar, R.~Girshick, K.~He, B.~Hariharan, S.~J. Belongie, Feature
      pyramid networks for object detection, in: Proceedings of IEEE Conference on
      Computer Vision and Pattern Recognition, 2017, pp. 936--944.
    
    \bibitem{cciccek20163d}
    {\"O}.~{\c{C}}i{\c{c}}ek, A.~Abdulkadir, S.~S. Lienkamp, T.~Brox,
      O.~Ronneberger, {3D} {U-Net}: Learning dense volumetric segmentation from
      sparse annotation, in: Proceedings of International Conference on Medical
      Image Computing and Computer Assisted Interventions, 2016, pp. 424--432.
    
    \bibitem{Zhao2017}
    H.~Zhao, J.~Shi, X.~Qi, X.~Wang, J.~Jia, Pyramid scene parsing network, in:
      Proceedings of IEEE Conference on Computer Vision and Pattern Recognition,
      2017, pp. 2881--2890.
    
    \bibitem{isensee2021nnu}
    F.~Isensee, P.~F. Jaeger, S.~A.~A. Kohl, J.~Petersen, K.~H. Maier-Hein,
      {nnU-Net}: A self-configuring method for deep learning-based biomedical image
      segmentation, Nature Methods 18~(2) (2021) 203--211.
    
    \bibitem{oktay2018attention}
    O.~Oktay, J.~Schlemper, L.~L. Folgoc, M.~Lee, M.~Heinrich, K.~Misawa, K.~Mori,
      S.~McDonagh, N.~Y. Hammerla, B.~Kainz, B.~Glocker, D.~Rueckert, {Attention
      U-Net}: Learning where to look for the pancreas (2018).
    \newblock \href {http://arxiv.org/abs/1804.03999} {\path{arXiv:1804.03999}}.
    
    \bibitem{huang2021missformer}
    X.~Huang, Z.~Deng, D.~Li, X.~Yuan, {MISSFormer}: An effective medical image
      segmentation transformer, arXiv preprint arXiv:2109.07162 (2021).
    
    \bibitem{Cao2023}
    H.~Cao, Y.~Wang, J.~Chen, D.~Jiang, X.~Zhang, Q.~Tian, M.~Wang, {Swin-Unet}:
      Unet-like pure transformer for~medical image segmentation, in: Proceedings of
      European Conference on Computer Vision Workshops, 2023, pp. 205--218.
    
    \bibitem{Chen2021}
    J.~Chen, Y.~Lu, Q.~Yu, X.~Luo, E.~Adeli, Y.~Wang, L.~Lu, A.~L. Yuille, Y.~Zhou,
      {TransUNet}: Transformers make strong encoders for medical image segmentation
      (2021).
    \newblock \href {http://arxiv.org/abs/2102.04306} {\path{arXiv:2102.04306}}.
    
    \bibitem{Hatamizadeh2022}
    A.~Hatamizadeh, Y.~Tang, V.~Nath, D.~Yang, A.~Myronenko, B.~Landman, H.~R.
      Roth, D.~Xu, {UNETR}: Transformers for 3d medical image segmentation, in:
      Proceedings of IEEE Winter Conference on Applications of Computer Vision,
      2022, pp. 574--584.
    
    \bibitem{shaker2023unetr}
    A.~Shaker, M.~Maaz, H.~Rasheed, S.~Khan, M.-H. Yang, F.~S. Khan, {UNETR++}:
      Delving into efficient and accurate 3d medical image segmentation (2023).
    \newblock \href {http://arxiv.org/abs/2212.04497} {\path{arXiv:2212.04497}}.
    
    \bibitem{Wilson2013}
    N.~M. Wilson, A.~K. Ortiz, A.~B. Johnson, The vascular model repository: A
      public resource of medical imaging data and blood flow simulation results,
      Journal of Medical Devices 7~(4) (2013) 0409231--409231.
    
    \bibitem{Radl2022}
    L.~Radl, Y.~Jin, A.~Pepe, J.~Li, C.~Gsaxner, F.-H. Zhao, J.~Egger, Avt:
      Multicenter aortic vessel tree cta dataset collection with ground truth
      segmentation masks, Data in Brief 40 (2022) 107801.
    
    \bibitem{Chen2018}
    L.-C. Chen, G.~Papandreou, I.~Kokkinos, K.~Murphy, A.~L. Yuille, Deeplab:
      Semantic image segmentation with deep convolutional nets, atrous convolution,
      and fully connected crfs, IEEE Transactions on Pattern Analysis and Machine
      Intelligence 40~(4) (2018) 834--848.
    
    \bibitem{vermeer2004model}
    K.~A. Vermeer, F.~M. Vos, H.~G. Lemij, A.~M. Vossepoel, A model based method
      for retinal blood vessel detection, Computers in Biology and Medicine 34~(3)
      (2004) 209--219.
    
    \bibitem{Thiagarajan2022}
    P.~Thiagarajan, P.~Khairnar, S.~Ghosh, Explanation and use of uncertainty
      quantified by bayesian neural network classifiers for breast histopathology
      images, IEEE Transactions on Medical Imaging 41~(4) (2022) 815--825.
    
    \bibitem{lorensen1998marching}
    W.~E. Lorensen, H.~E. Cline, Marching cubes: A high resolution 3d surface
      construction algorithm, in: Seminal graphics: pioneering efforts that shaped
      the field, ACM, 1998, pp. 347--353.
    
    \bibitem{valette2004approximated}
    S.~Valette, J.-M. Chassery, Approximated centroidal voronoi diagrams for
      uniform polygonal mesh coarsening, Computer Graphics Forum 23~(3) (2004)
      381--389.
    
    \bibitem{ravi2020accelerating}
    N.~Ravi, J.~Reizenstein, D.~Novotny, T.~Gordon, W.-Y. Lo, J.~Johnson,
      G.~Gkioxari, Accelerating 3d deep learning with pytorch3d, arXiv preprint
      arXiv:2007.08501 (2020).
    
    \bibitem{qi2017pointnet++}
    C.~R. Qi, L.~Yi, H.~Su, L.~J. Guibas, Pointnet++: Deep hierarchical feature
      learning on point sets in a metric space, Advances in neural information
      processing systems 30 (2017).
    
    \bibitem{nealen2006laplacian}
    A.~Nealen, T.~Igarashi, O.~Sorkine, M.~Alexa, Laplacian mesh optimization, in:
      Proceedings of the 4th international conference on Computer graphics and
      interactive techniques in Australasia and Southeast Asia, 2006, pp. 381--389.
    
    \end{thebibliography}

\end{document}